\documentclass[journal]{IEEEtran}
\usepackage{ifpdf}

\usepackage{cite}

\ifCLASSINFOpdf
  \usepackage[pdftex]{graphicx}
\else
  \usepackage[dvips]{graphicx}
\fi
\usepackage{amsmath, amssymb}
\usepackage{algorithmic, algorithm}

\usepackage{array}

\ifCLASSOPTIONcompsoc
 \usepackage[caption=false,font=normalsize,labelfont=sf,textfont=sf]{subfig}
\else
 \usepackage[caption=false,font=footnotesize]{subfig}
\fi

\usepackage{stfloats}
\ifCLASSOPTIONcaptionsoff
 \usepackage[nomarkers]{endfloat}
\let\MYoriglatexcaption\caption
\renewcommand{\caption}[2][\relax]{\MYoriglatexcaption[#2]{#2}}
\fi
\usepackage{url}

\usepackage[colorlinks,linkcolor=red,anchorcolor=green,citecolor=blue]{hyperref}

\usepackage{multirow}
\usepackage[table,xcdraw]{xcolor}
\usepackage{longtable}
\usepackage{subfig}
\usepackage{booktabs}
\usepackage{threeparttable}
\usepackage{lscape}
\usepackage{hyperref}
\usepackage{pifont}

\newif\ifarxiv
\arxivtrue
\usepackage{lastpage}
\usepackage{fancyhdr}
\fancypagestyle{firstpage}
{
	\fancyhead{}
	
	\ifarxiv
	\fancyfoot[C]{\scriptsize \copyright~2023 IEEE. Personal use of this material is permitted. Permission from IEEE must be obtained for all other uses, in any current or future media, including reprinting/republishing this material for advertising or promotional purposes, creating new collective works, for resale or redistribution to servers or lists, or reuse of any copyrighted component of this work in other works. 
		This work has been accepted by the IEEE Transactions on Computer-Aided Design of Integrated Circuits and Systems (TCAD).}
	\else
	\fi
}

\begin{document}

\newcommand{\rOne}[1]{\textcolor{black}{#1}}
\newcommand{\rTwo}[1]{\textcolor{black}{#1}}
\newcommand{\rThree}[1]{\textcolor{black}{#1}}

%
\title{Efficient \textit{N:M} Sparse DNN Training Using \\ Algorithm, Architecture, and Dataflow Co-Design}
%
%

%

\author{Chao~Fang,~\IEEEmembership{Graduate~Student~Member,~IEEE,}
        Wei~Sun,
        Aojun~Zhou, 
        and~Zhongfeng~Wang,~\IEEEmembership{Fellow,~IEEE}
\thanks{This work was supported in part by the National Key R\&D Program of China under Grant 2022YFB4400604 and in part by the National Natural Science Foundation of China under Grant 62174084. \textit{(Corresponding author: Zhongfeng Wang.)}}
\thanks{C. Fang and Z. Wang are with the School of Electronic Science and Engineering, Nanjing University, Nanjing 210008, China (e-mail: fantasysee@smail.nju.edu.cn; zfwang@nju.edu.cn).}
\thanks{W. Sun is with Electronic System Group, Eindhoven University of Technology, Netherlands (e-mail: w.sun@tue.nl).}
\thanks{A. Zhou is with CUHK-Sensetime Joint Lab, CUHK, Hong Kong, China (e-mail: aojunzhou@link.cuhk.edu.hk).}
}

%
%

\markboth{To appear in the IEEE Transactions on Computer-Aided Design of Integrated Circuits and Systems}%
{Shell \MakeLowercase{\textit{et al.}}: Bare Demo of IEEEtran.cls for IEEE Journals}
%



\maketitle
\thispagestyle{firstpage}

\begin{abstract}
\rOne{Sparse training is one of the promising techniques to reduce the computational cost of DNNs while retaining high accuracy. 
In particular, \textit{N:M} fine-grained structured sparsity, where only \textit{N} out of consecutive \textit{M} elements can be nonzero, has attracted attention due to its hardware-friendly pattern and capability of achieving a high sparse ratio.}
However, the potential to accelerate \textit{N:M} sparse DNN training has not been fully exploited, and there is a lack of efficient hardware supporting \textit{N:M} sparse training.
To tackle these challenges, this paper presents a computation-efficient training scheme for \textit{N:M} sparse DNNs using algorithm, architecture, and dataflow co-design.
At the algorithm level, a bidirectional weight pruning method, dubbed BDWP, is proposed to leverage the \textit{N:M} sparsity of weights during both forward and backward passes of DNN training, which can significantly reduce the computational cost while maintaining model accuracy.
At the architecture level, a sparse accelerator for DNN training, namely SAT, is developed to neatly support both the regular dense operations and the computation-efficient \textit{N:M} sparse operations.
At the dataflow level, multiple optimization methods ranging from interleave mapping, pre-generation of \textit{N:M} sparse weights, and offline scheduling, are proposed to boost the computational efficiency of SAT.
Finally, the effectiveness of our training scheme is evaluated on a Xilinx VCU1525 FPGA card using various DNN models \rTwo{(ResNet9, ViT, VGG19, ResNet18, and ResNet50)} and datasets \rTwo{(CIFAR-10, CIFAR-100, Tiny ImageNet, and ImageNet)}.
\rTwo{Experimental results show the SAT accelerator with the BDWP sparse training method under 2:8 sparse ratio achieves an average speedup of 1.75$\times$ over that with the dense training, accompanied by a negligible accuracy loss of 0.56\% on average. Furthermore, our proposed training scheme significantly improves the training throughput by 2.97$\sim$25.22$\times$ and the energy efficiency by 1.36$\sim$3.58$\times$ over prior FPGA-based accelerators.}
\end{abstract}

%
\IEEEpeerreviewmaketitle

\section{Introduction}
\IEEEPARstart{D}{eep} neural networks (DNNs) have been widely used in many applications, such as computer vision, speech recognition, autonomous driving, and robotics.
However, their impressive accuracy comes at the cost of expensive computational requirements: the state-of-the-art DNNs could contain trillions of parameters \cite{fedus2022switch} and consume thousands of peta-level floating-point operations (FLOPs) \cite{brown2020language} for the training process.
To relieve this burden, it is crucial to seek a computation-efficient scheme for DNN training.

Sparse training \cite{mcdanel2022accelerating, zhang2022xst, song2019approximate, yuan2021growing, zhang2021structadmm, ye2020accelerating, evci2020rigging, wu2020enabling, hassantabar2021scann, park2022quantized} is one of the promising techniques to reduce the training computational cost while retaining impressive accuracy of DNNs. 
It dynamically prunes elements such as weights, activations, gradients, and then eliminates computations associated with the pruned elements during training iterations.
Prior works mostly focus on exploiting structured \cite{yuan2021growing, zhang2021structadmm} or unstructured \cite{evci2020rigging, ye2020accelerating, wu2020enabling, hassantabar2021scann} sparsity patterns for sparse DNN training.
Structured sparsity \cite{yuan2021tinyadc, liang2020omni, tu2022sdp}, which involves pruning entire kernels or channels of weights at a coarse-grained level, has limitations in reducing the number of FLOPs of DNN training (less than 40\% sparse ratio in \cite{yuan2021growing}).
On the other hand, unstructured sparsity \cite{li2020fsa, liu2021search, jung2022energy} involves pruning elements in any position without constraints, leading to a high sparse ratio (over 80\% sparse ratio in \cite{evci2020rigging}) and a significant reduction in the number of FLOPs for training.
However, its irregular pattern makes it challenging to be fully utilized on hardware for effective training speedup \cite{wang2019low, a100, mishra2021accelerating}.

In recent years, there has been a growing interest in leveraging fine-grained structured sparsity \cite{kundu2020pre, niu2020patdnn, mishra2021accelerating, ozen2022unleashing, huang2022structured} for efficient DNN acceleration.
\textit{N:M} sparsity \cite{mishra2021accelerating} has attracted significant attention among various fine-grained structured sparsity for its practical sparsity ratio and hardware-friendly pattern, which allows only \textit{N} out of consecutive \textit{M} elements in a group to be nonzero.
The static 2:4 sparse pattern was initially introduced by NVIDIA Ampere GPUs \cite{a100} for efficient DNN inference.
However, researchers have gone beyond the 2:4 static pattern and explored more aggressive \textit{N:M} sparse patterns, such as 2:8 or 2:16, which have demonstrated significant inference acceleration \cite{xie2023efficient, fang2022algorithm, chen2023dynamic} with competitive accuracy to the dense counterparts \cite{lin20221xn, frantar2022spdy, sun2021dominosearch}.
In addition to its use in efficient DNN inference, \textit{N:M} sparsity has also shown potential in reducing the computational cost of DNN training \cite{mcdanel2022accelerating, zhou2021learning}.
However, accelerating \textit{N:M} sparse DNN training is a challenging task that involves several issues to be addressed. 
\begin{itemize}
    \item The potential to accelerate \textit{N:M} sparse DNN training has not been fully exploited. 
    Previous works solely accelerate DNN training by introducing \textit{N:M} sparsity in either forward pass \cite{zhou2021learning} or backward pass \cite{mcdanel2022accelerating}. 
    A unified approach that takes advantage of \textit{N:M} sparsity in both passes could lead to further acceleration of DNN training.
    \item Current hardware platforms are unable to fully leverage the sparsity benefits of \textit{N:M} sparsity to accelerate DNN training.
    Firstly, the Ampere GPUs \cite{a100} only support 2:4 sparse operations, which restricts the acceleration capability of various \textit{N:M} patterns with higher sparsity ratios for efficient DNN training. 
    Secondly, while \textit{N:M} sparse data can be packed in advance for DNN inference deployment, the \textit{N:M} elements need to be updated in training iterations. 
    However, the lack of dedicated hardware implementation for this iterative update process results in substantial computational overhead, hampering the speedup of \textit{N:M} sparse training.
    \item \rOne{Dataflow mapping optimizations are required to further improve hardware utilization and achieve significant acceleration. Specifically, in \textit{N:M} sparse DNN training, various types of computational intensive operations in both forward and backward passes, require tailored dataflow optimizations for both dense and \textit{N:M} sparse operations to better utilize hardware resources and accelerate the training process.}
\end{itemize}

To address the aforementioned issues, this paper presents a computation-efficient \textit{N:M} sparse training scheme for DNNs, featuring three aspects: algorithm, architecture, and dataflow.
\textbf{1)} The bidirectional weight pruning algorithm for \textit{N:M} sparse training, namely BDWP, leverages the \textit{N:M} sparse pattern on weights in both forward and backward passes, and significantly reduces the number of training operations.
\textbf{2)} The efficient hardware architecture for \textit{N:M} sparse DNN training, dubbed as SAT, supports both regular dense matrix multiplication (MatMul) and \textit{N:M} sparse MatMul with improved training throughput and energy efficiency. Additionally, SAT also supports online \textit{N:M} sparse reduction of data, further enhancing its computational efficiency. 
\textbf{3)} The dataflow optimization techniques, including interleave mapping, pre-generation of \textit{N:M} sparse weights, and offline scheduling, further increase hardware utilization and improve the throughput of SAT.

The main contributions can be summarized as follows:

\begin{itemize}
    \item \textbf{Algorithm}: We propose a bidirectional weight pruning method for \textit{N:M} sparse training, namely BDWP, leveraging \textit{N:M} sparsity of weights during both forward and backward passes of DNN training. 
    \rOne{Compared to dense training, our 2:8 BDWP training reduces the number of training operations by 48\% with a negligible accuracy loss of \rTwo{0.56\%} on average.}
    \item \textbf{Architecture}: We propose a sparse accelerator for training DNNs, namely SAT, to support computation-efficient \textit{N:M} sparse operations besides the regular dense operations. It achieves 2.97$\sim$25.22$\times$ higher throughput and 1.36$\sim$3.58$\times$ greater energy efficiency than prior training accelerators \cite{venkataramanaiah2020fpga, tang2022ef, he2020fecaffe, liu2017fpga, venkataramanaiah2019automatic, chen2021eile, tsai2023chip} evaluated on FPGA.
    \item \textbf{Dataflow}: We propose several dataflow optimization methods, including interleave mapping, pre-generation of \textit{N:M} sparse weights, and offline scheduling, to boost the computational efficiency of SAT.
    \item \textbf{Scheme}: We present an efficient scheme for DNN training that incorporates the BDWP algorithm and the SAT architecture. It improves the training speed by \rTwo{1.75$\times$} on average compared to the conventional dense training scheme deployed on SAT.
\end{itemize}
 
The remainder of this paper is organized as follows. Sec.~\ref{sec:bkg} introduces training steps of DNNs and reviews sparse training techniques and FPGA-based training accelerators. 
Sec.~\ref{sec:algo}, Sec.~\ref{sec:arch}, and Sec.~\ref{sec:dataflow} describe BDWP algorithm, SAT architecture, and optimization methods for SAT dataflow, respectively.
Sec.~\ref{sec:res} presents experimental results to illustrate the effectiveness of our proposed computation-efficient scheme for \textit{N:M} sparse DNN training. 
The paper is concluded in Sec.~\ref{sec:concls}.
\begin{figure} [htbp] 
	\centering
	\includegraphics[width=0.50\textwidth]{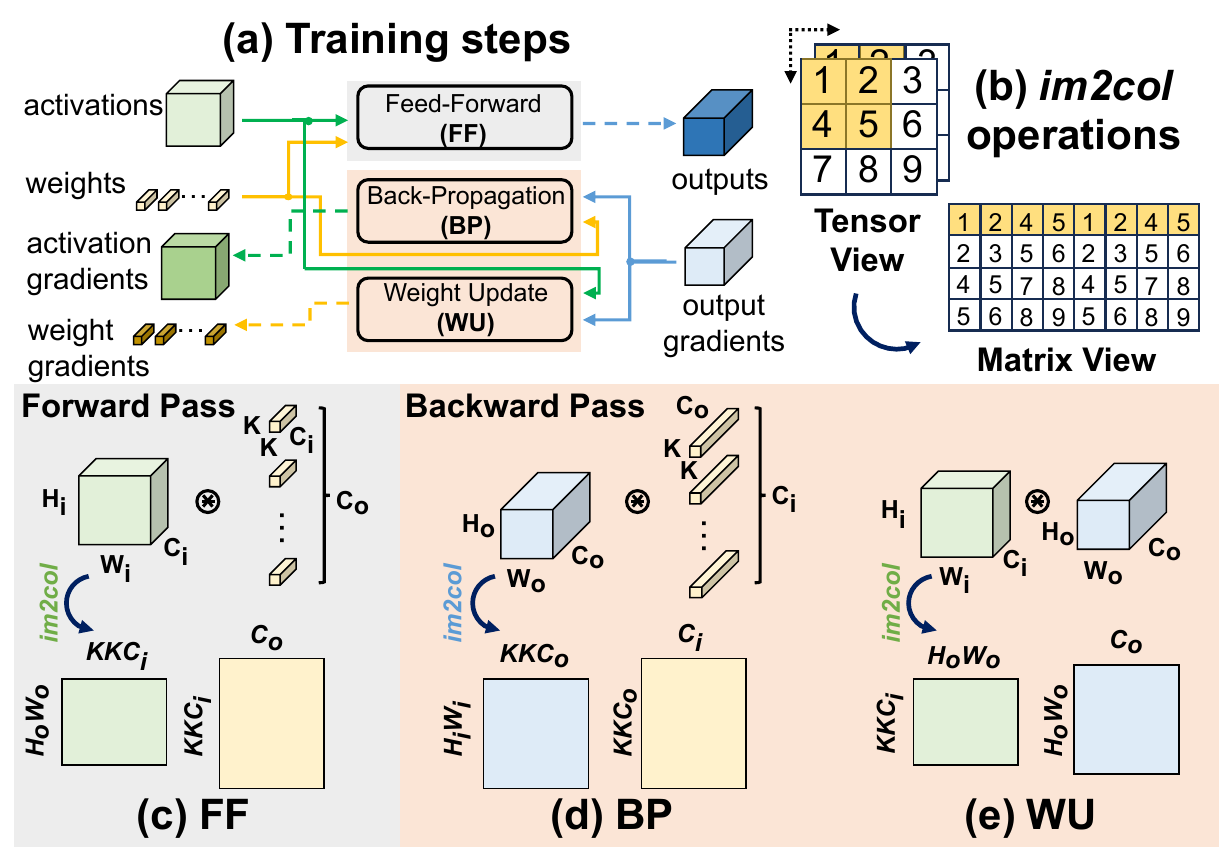}
	\caption{\rTwo{Example of the training process for a convolutional layer: (a) it consists of FF in the forward pass, BP, and WU in the backward pass. An input tensor can be transformed into a matrix if required by the (b) \textit{im2col} process. (c), (d), and (e) illustrate the MatMul operations during FF, BP, and WU, respectively, using a single batch data.}}
	\label{fig:train_steps}
\end{figure}

\section{Background and Related Works} \label{sec:bkg}

\subsection{Training Steps of DNNs}

\begin{figure} [htbp] 
	\centering
	\includegraphics[width=0.48\textwidth]{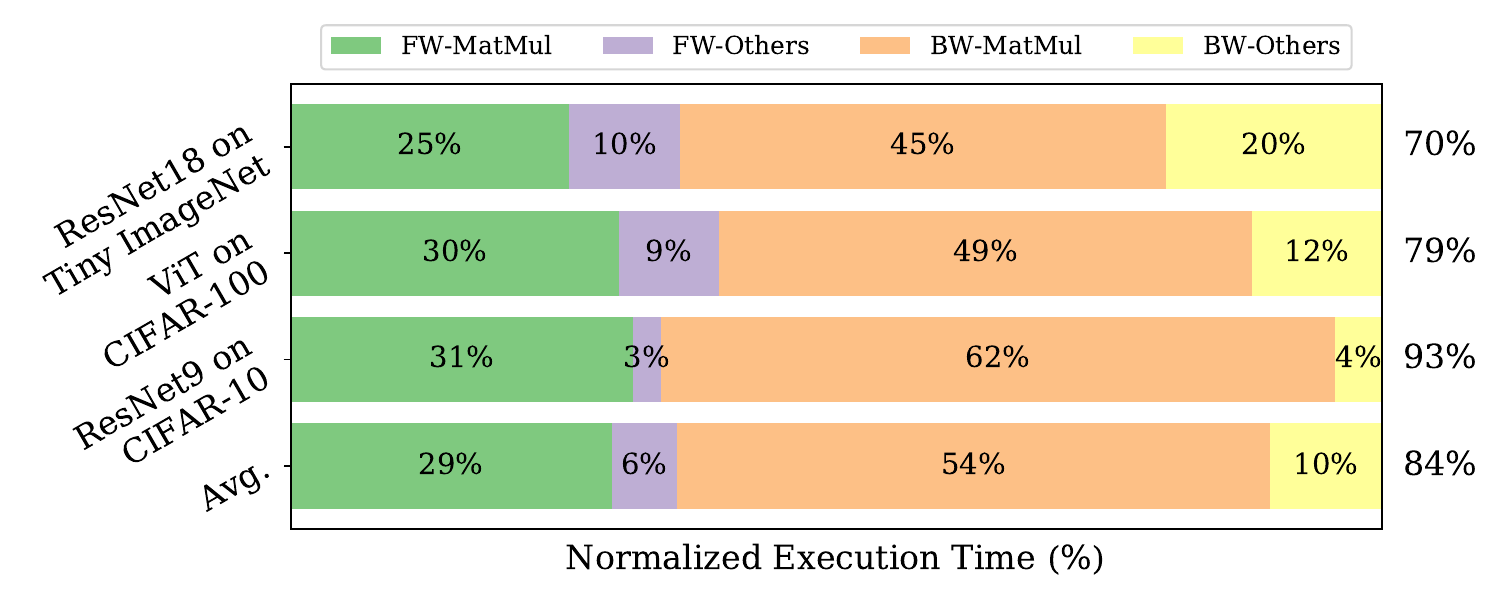}
	\caption{Execution time profile results on training three DNN models with a batch size of 512 using an RTX 2080 Ti card. The dominance of MatMul operations in the training process highlights the significance of accelerating these operations to improve training efficiency.}
	\label{fig:model_profile}
\end{figure}

As shown in Fig.~\ref{fig:train_steps}~(a), the training steps of a DNN layer consist of two stages: the forward and the backward passes.
In the forward pass, the network layer feeds forward (FF) the activations as input and produces the outputs based on the trainable weights.
The intermediate outputs computed during FF must be kept in memory for the backward pass.
In the backward pass, the gradients of activations and weights are computed by backward propagation (BP) and weight update (WU), respectively.
The activation gradients are calculated based on the weights and the output gradients, while the weight gradients are computed according to the stored activations and the output gradients.
MatMul can be utilized to unify the computational ﬂow of computationally intensive layers like convolutional layers of ResNet18 and linear layers of ViT in large-batch DNN training, boosting computational efficiency. 
\rTwo{Taking a convolutional layer as an example, Fig.~\ref{fig:train_steps}~(b) demonstrates the \textit{im2col} process \cite{rohwedder2021pooling}, which converts a tensor into a matrix for subsequent DNN training operations.
Fig.~\ref{fig:train_steps}~(c) to (e) further illustrates how FF, BP, and WU for a single layer of DNN are transformed into MatMul through \textit{im2col} process, respectively.}
Furthermore, we leverage PyTorch profiler \cite{torch_profiler} to dissect the execution time for training three typical DNNs with a batch size of 512 using an RTX 2080 Ti card.
As shown in Fig.~\ref{fig:model_profile}, those operations can be unified into MatMuls and constitute a considerable portion, up to 84\%, of the training time per batch on average, significantly impacting the training process.
\rTwo{By optimizing these MatMuls, significant improvements in training speed can be attained.}

\subsection{Sparse Training for DNNs}

Sparse training \cite{mcdanel2022accelerating, zhang2022xst, song2019approximate, yuan2021growing, zhang2021structadmm, ye2020accelerating, evci2020rigging, wu2020enabling, hassantabar2021scann, park2022quantized} is one of the promising techniques to reduce the training computational cost while retaining impressive accuracy of DNNs.
A majority of prior arts mainly focus on \rTwo{ReLU-based \cite{dai2020sparsetrain, lu2022theta}}, structured \cite{yuan2021growing, zhang2021structadmm} or unstructured sparsity \cite{evci2020rigging, ye2020accelerating} for sparse training.
\rTwo{ReLU-based sparsity has no impact on model accuracy but suffers a low sparse ratio (approximately 50\%) and can be difficult to be exploited on hardware.}
Structured sparsity has limitations in reducing computational complexity (less than 40\% sparse ratio in \cite{yuan2021growing}), while unstructured sparsity, despite saving a great number of operations (over 80\% sparse ratio in \cite{evci2020rigging}), is difficult to accelerate on hardware \cite{a100}.
To achieve a better trade-off between structured and unstructured sparsity, pre-defined sparsity \cite{kundu2020pre, niu2020patdnn}, has been exploited as the pioneering technique, where the sparse pattern is predetermined based on prior knowledge or heuristics.
Recently, \textit{N:M} fine-grained structured sparsity \cite{mishra2021accelerating}, \rTwo{demanding} that at least \textit{N} values must be zero for each group of \textit{M} values, has attracted the attention of researchers \cite{zhou2021learning, sun2021dominosearch, mcdanel2022accelerating, lin20221xn, frantar2022spdy, fang2022algorithm, chen2023dynamic, xie2023efficient, fang2023cest} due to its practical sparsity ratio as well as its hardware-friendly pattern.
As for \textit{N:M} fine-grained sparse training, \rTwo{SR-STE} \cite{zhou2021learning} and SDGP \cite{mcdanel2022accelerating} can boost training efficiency by forcing \textit{N:M} sparsity on weights in the forward pass and output gradients in the backward pass, respectively.
The sparsity introduced in one training direction hinders the further improvement of training efficiency.
In this work, our BDWP leverages the \textit{N:M} sparsity of weights in both forward and backward passes of DNN training, significantly reducing the number of operations and retaining competitive training accuracy compared to \rTwo{SR-STE} and SDGP.

\subsection{FPGA-based DNN Training Accelerators}

Nowadays, DNN training is mainly accelerated on power-hungry GPU devices \cite{yuan2021growing, mcdanel2022accelerating} leading to low energy efficiency \cite{venkataramanaiah2019automatic}.
Additionally, many ASIC designs dedicated to DNN training, such as \cite{wang2022trainer}, achieve good energy efficiency but require lengthy design cycles.
In contrast, building dedicated accelerators on FPGA enable agile deployment with satisfactory energy efficiency.
Various DNN training accelerators \cite{venkataramanaiah2019automatic, venkataramanaiah2020fpga, tang2022ef, liu2017fpga, shao2021fpga, luo2020towards, lu2022eta, nakahara2019fpga, chen2021eile, tsai2023chip} have been developed on FPGA platforms.
\cite{venkataramanaiah2019automatic, venkataramanaiah2020fpga, tang2022ef, liu2017fpga} exploited optimization for standard DNN training process.
However, with the increasing number of operations in developing DNNs, it is difficult to achieve satisfying speedup by simply optimizing dataflow or hardware design for the standard training process.
\rOne{Furthermore, \cite{nakahara2019fpga, tsai2023chip} exploit DNN sparse training acceleration on FPGA but suffer low computational efficiency due to leveraging irregular unstructured sparse patterns.
In addition, \cite{shao2021fpga, luo2020towards, lu2022eta, nakahara2019fpga} employed aggressive reduced numerical precision, such as FP9 or INT8, to decrease the computational cost of DNN training which is orthogonal to sparse DNN training.}
\rOne{Compared to prior works, we aim to significantly improve DNN training efficiency by developing an FPGA-based accelerator, namely SAT, that enables both dense and computation-efficient \textit{N:M} sparse operations within the \textit{N:M} sparse DNN training process.}

\begin{figure} [htbp] 
	\centering
	\includegraphics[width=0.46\textwidth]{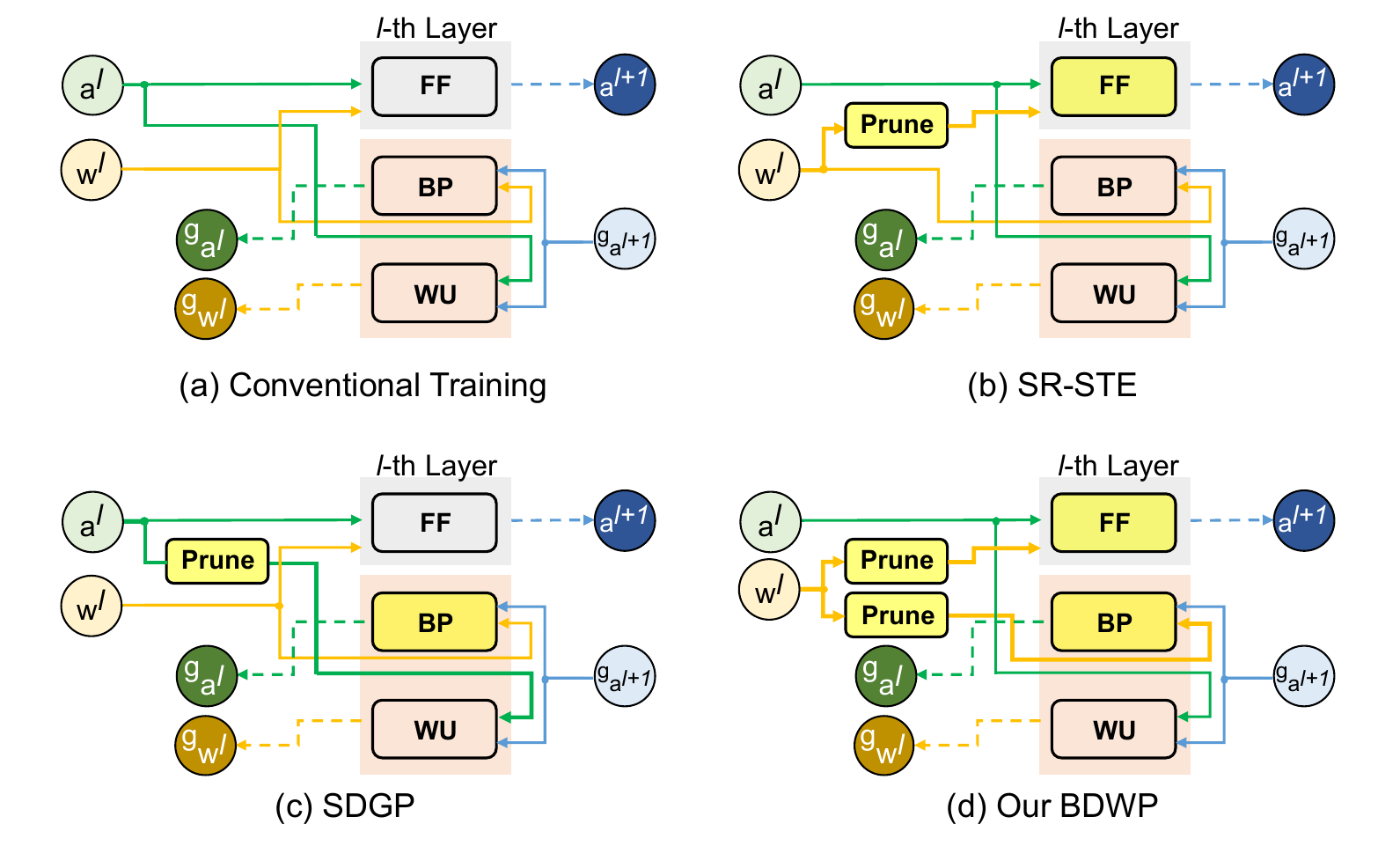}
	\caption{Comparison of (a) conventional training, uni-directional \textit{N:M} sparse training, including (b) \rTwo{SR-STE} and (c) SDGP, and our proposed bidirectional \textit{N:M} sparse training, i.e., (d) BDWP for a DNN layer. Without compromising convergent accuracy, the DNN training process using BDWP can significantly speed up due to aggressive \textit{N:M} pruning in both forward and backward passes.} 
	\label{fig:algo_cmp}
    \vspace{-1em}
\end{figure}

\section{\textit{N:M} Sparse Training Algorithm} \label{sec:algo}

\rOne{\textit{N:M} sparsity pattern can be leveraged to accelerate the DNN training process by significantly reducing the number of operations, which has been introduced through \rTwo{SR-STE} \cite{zhou2021learning} in the forward pass and SDGP \cite{mcdanel2022accelerating} in the backward pass.
In this section, we first learn the sensitivity of the training loss by introducing \textit{N:M} sparse patterns during DNN training and then propose our BDWP based on our findings, which unifies \textit{N:M} patterns in both forward and backward passes to further elevate DNN training efficiency.
}

\begin{figure} [tbp] 
	\centering
	\includegraphics[width=0.48\textwidth]{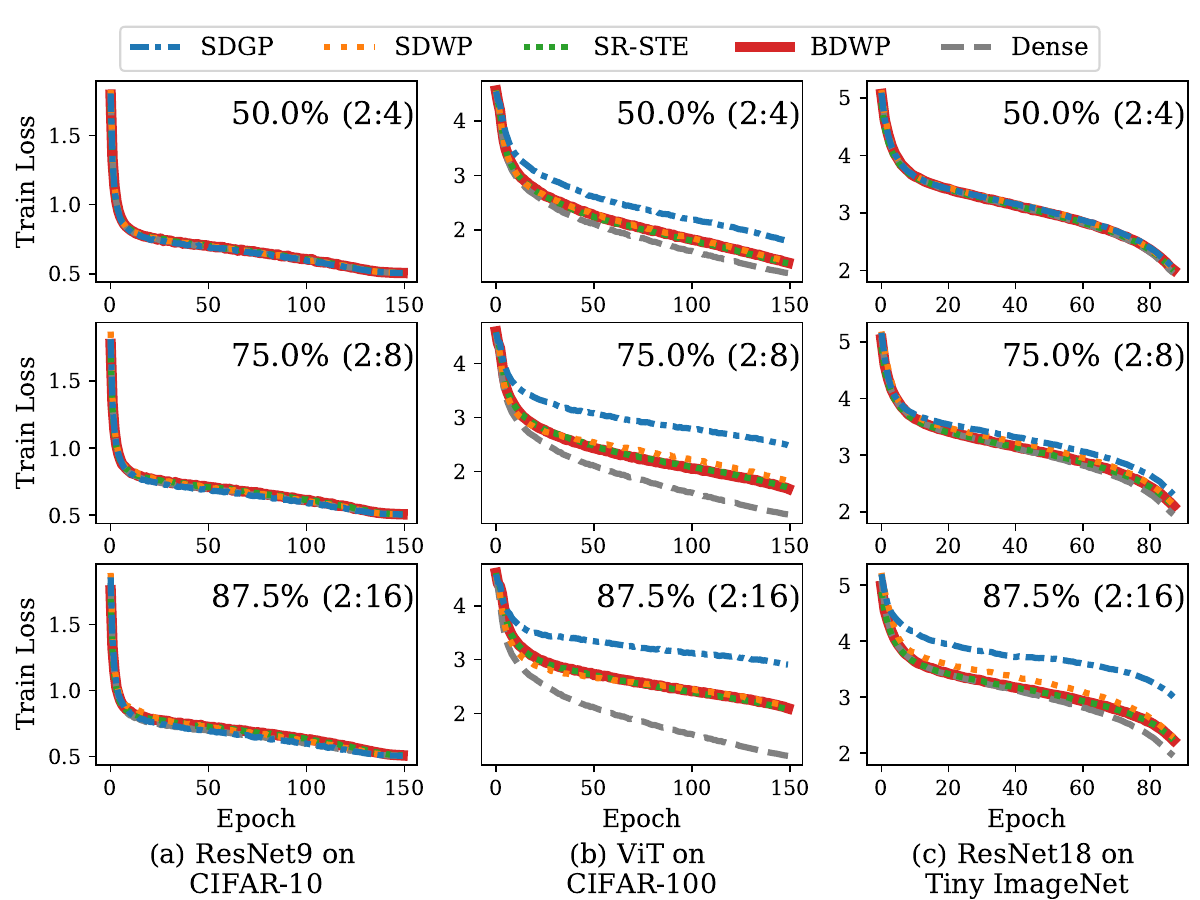}
    \caption{\rTwo{Comparison of training loss using multiple \textit{N:M} pruning methods for (a) ResNet9 on CIFAR-10, (b) ViT on CIFAR-100 dataset and (c) ResNet18 on Tiny ImageNet dataset.}}
	\label{fig:loss}
\end{figure}

\subsection{Exploiting \textit{N:M} Sparse Training Potential}
\rOne{As shown in Fig.~\ref{fig:algo_cmp}, compared with the conventional training, \rTwo{SR-STE} prunes weights in the forward pass, while SDGP prunes output gradients in the backward pass to reduce the number of required operations of DNNs.}
\rThree{To evaluate the effectiveness of these \textit{N:M} pruning techniques, we employ the from-scratch training loss as a metric and compare the errors of the pruned models with those of the densely trained models.}
\rTwo{Fig.~\ref{fig:loss} presents the loss curves when training from scratch ResNet9, ViT, and ResNet18 models on CIFAR-10, CIFAR-100, and Tiny ImageNet datasets, respectively.
For ResNet9 on the simple CIFAR-10 dataset, both SR-STE and SDGP exhibit good performance compared to the dense baseline.
However, for larger models or more complicated datasets like ViT on CIFAR-100 and ResNet18 on Tiny ImageNet, pruning activations in the backward pass using SDGP with a sparse ratio of about 75\% results in a loss curve that deviates significantly from the dense training scheme.
To address this issue, we explore an alternative approach by pruning weights in the backward pass, denoted as SDWP in Fig.~\ref{fig:loss}. 
Notably, SDWP demonstrates better convergence compared to SDGP at the same \textit{N:M} sparse ratio.
Therefore, a novel bidirectional \textit{N:M} sparse training approach called BDWP is proposed by integrating both unidirectional weight pruning techniques, i.e., \rTwo{SR-STE} and SDWP.
The training loss curve of BDWP in Fig.~\ref{fig:loss} closely aligns with \rTwo{SR-STE} and SDWP at the same \textit{N:M} sparse ratio, showing BDWP achieving negligible impact on training convergence with a significant reduction of training operations.}

\begin{algorithm}
	\renewcommand{\algorithmicrequire}{\textbf{Input:}}
	\renewcommand{\algorithmicensure}{\textbf{Output:}}
	\caption{Training an $L$ layer network using BDWP}
	\label{alg:bdwp}
	\begin{algorithmic}[1]
	    \REQUIRE A mini-batch of input activations and labels ($a^0_t$, $y_t$), current weights $w_t$, sparse ratio $N$ and $M$ at iteration $t$.
	    \ENSURE Updated weights $w_{t+1}$. \\
	    \textbf{Forward Pass} \\
        \FOR{$l = 1$ to $L$}
            \STATE $ \widetilde{w}^{l}_{\text{FF}} \leftarrow $ BDWP$_{\text{FF}}$($w^l_t$, $N$, $M$).
            \STATE $a^l_t \leftarrow $ \textbf{FF}($a^{l-1}_t$, $\widetilde{w}^{l}_{\text{FF}}$). 
        \ENDFOR 
        \STATE Compute the gradient of the output layer $g_{a^{L}_{t}}$. \\
        \textbf{Backward Pass} \\
        \FOR{$l = L$ downto $1$}
            \STATE $ \widetilde{w}^{l}_{\text{BP}} \leftarrow $ BDWP$_{\text{BP}}$($w^l_t$, $N$, $M$).
            \STATE $g_{a^{l-1}_{t}} \leftarrow $ \textbf{BP}($g_{a^{l}_{t}}$, $\widetilde{w}^{l}_{\text{BP}}$).
            \STATE $g_{w^{l-1}_{t}} \leftarrow $ \textbf{WU}($a^{l-1}_t$, $g_{a^{l}_{t}}$).
        \ENDFOR
        \STATE Optimize $w_{t+1}$ with momentum SGD.
	\end{algorithmic} 
\end{algorithm} 

\subsection{Bidirectional Weight Pruning}

\begin{figure} [tbp] 
	\centering
	\includegraphics[width=0.46\textwidth]{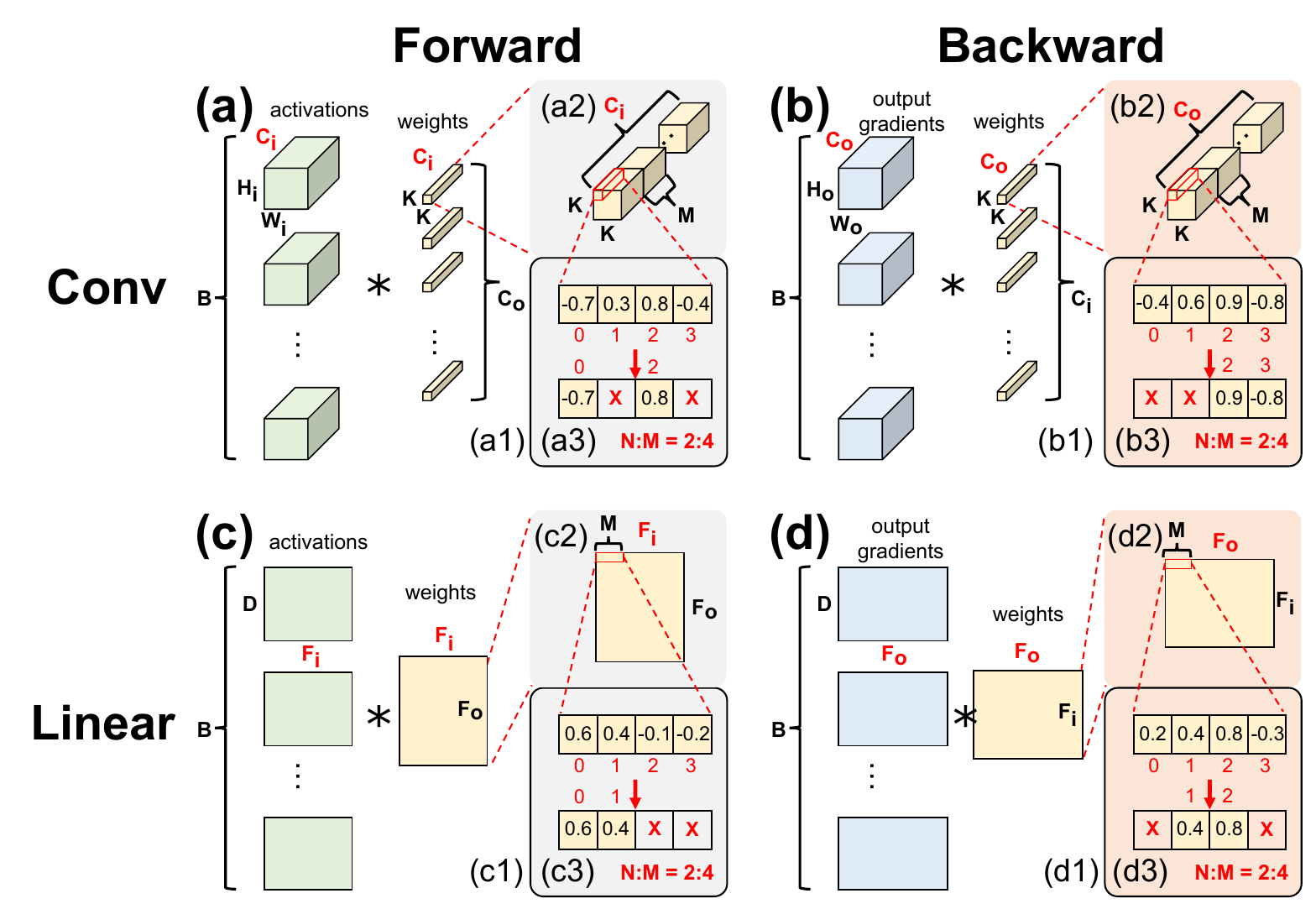}
	\caption{Applying BDWP in the convolutional layer and the linear layer in both forward and backward passes of DNN training. For the convolutional layer, BDWP is applied to each group across (a) input channels in the forward pass and (b) output channels in the backward pass, respectively. For the linear layer, BDWP is applied to each group across (c) input features in the forward pass and (d) output features in the backward pass, respectively.} 
	\label{fig:prune}
\end{figure}

Our \textit{N:M} sparse training method, BDWP, is illustrated in Fig.~\ref{fig:algo_cmp}~(d) and detailed in Algorithm~\ref{alg:bdwp}.

\textbf{Notation.} Given a mini-batch of training samples $a^0_t$ and labels $y_t$, we aim to optimize weights $w_t$ to $w_{t+1}$ at iteration $t$ for an $L$ layer DNN.
Activations and weights of the $l$-th layer are denoted as $a^l_t$ and $w^l_t$, respectively.
In addition, $g_{a^{l}_{t}}$ and $g_{w^{l}_{t}}$ denote gradients with respect to activations and weights of the $l$-th layer, respectively.
Weights of the corresponding sparse network are denoted with $\widetilde{w}$, and $\widetilde{w}_{\text{FF}}$ and $\widetilde{w}_{\text{BP}}$ denote the pruned \textit{N:M} sparse weights of BDWP using in the forward pass and backward pass, respectively.

\textbf{Training Flow.} 
Algorithm~\ref{alg:bdwp} describes the process of BDWP when training an $L$ layer network at the iteration $t$.
BDWP$_{\text{FF}}$ and BDWP$_{\text{BP}}$ denote the \textit{N:M} element group generation process in the forward pass and backward pass, respectively.
Both of them take the dense weights $w$ as input and generate \textit{N:M} sparse weights $\widetilde{w}$ for further computation.
In FF, as shown in Line 3, the activations perform operations with the \textit{N:M} sparse $\widetilde{w}_{\text{FF}}$, which have been slimmed for $\frac{M}{N}$ times.
In BP, as shown in Line 8, the activation gradients are obtained by performing operations with output gradients and $\widetilde{w}_{\text{BP}}$.
The other steps of the training process stay the same as the standard training flow.
\rOne{For from-scratch training, \textit{N:M} sparse patterns are leveraged from the first to the final training steps, and following \rTwo{SR-STE} and SDGP, updated in every training step. Other training hyperparameters remain the same as for dense training.}

\rTwo{\textbf{\textit{N:M} Element Group Generation.} 
Fig.~\ref{fig:prune} presents how to apply BDWP in forward and backward passes to the convolutional layer and the linear layer, respectively, both of which dominate most of the required DNN training operations.
BDWP preserves values with the $N$ most significant magnitude in each group of $M$ elements.
\textit{N:M} is assumed as 2:4 here, and $B$ denotes batch size. Additionally, the subscripts $i$ and $o$ refer to the input activations and output gradients, respectively.
When training a convolutional layer, the input activations are represented by a tensor with $B$ batches, each having a height of $H$, width of $W$, and $C_i$ channels, and the weights are denoted as a tensor with a height of $K$, width of $K$, $C_i$ input channels, and $C_o$ output channels.
BDWP removes pruned elements in each group across the input channels ($C_i$) in the forward pass, and across the output channels ($C_o$) in the backward pass, respectively.
As for training a linear layer, the input activations are represented by a tensor with $B$ batches, each having a width of $D$ and $F_i$ features, and the weights are represented by a transformation matrix from $F_i$ to $F_o$.
BDWP is applied to each group across input features ($F_i$) in the forward pass and across output features ($F_o$) in the backward pass, respectively.
The preserved elements in each group would be packed into the compact format as in \cite{mishra2021accelerating} to reduce memory consumption.}

\textbf{Opportunities on Hardware Implementation.}
\rOne{
Given hardware accelerators supporting \textit{N:M} sparsity are rather limited \cite{a100, fang2022algorithm}, there are opportunities to develop new architectures to accelerate \textit{N:M} sparse training with high performance and efficiency.
For instance, generating \textit{N:M} sparse weights during each iteration presents an opportunity for an on-chip hardware module capable of producing \textit{N:M} sparse data.
In addition, to accommodate the varying sizes of MatMuls required at different stages of the training process, a flexible interconnect capable of adapting to the changing demands is imperative.
Finally, a unified and efficient computing unit is critical to enable the accelerator to handle both \textit{N:M} sparse and dense operations with high computational efficiency.
}

\section{Hardware Architecture} \label{sec:arch}

\rOne{
The efficient hardware architecture is crucial for achieving significant acceleration in DNN training through computational optimization resulting from \textit{N:M} sparsity pattern.
This section presents an efficient \textit{N:M} sparse accelerator for DNN training, namely SAT, fully leveraging the computation-efficient operations from \textit{N:M} sparse training algorithms.
We first briefly introduce the overall architecture of SAT and then elaborate on the designs of the major computing engines.
}

\begin{figure} [hbtp] 
	\centering
	\includegraphics[width=0.46\textwidth]{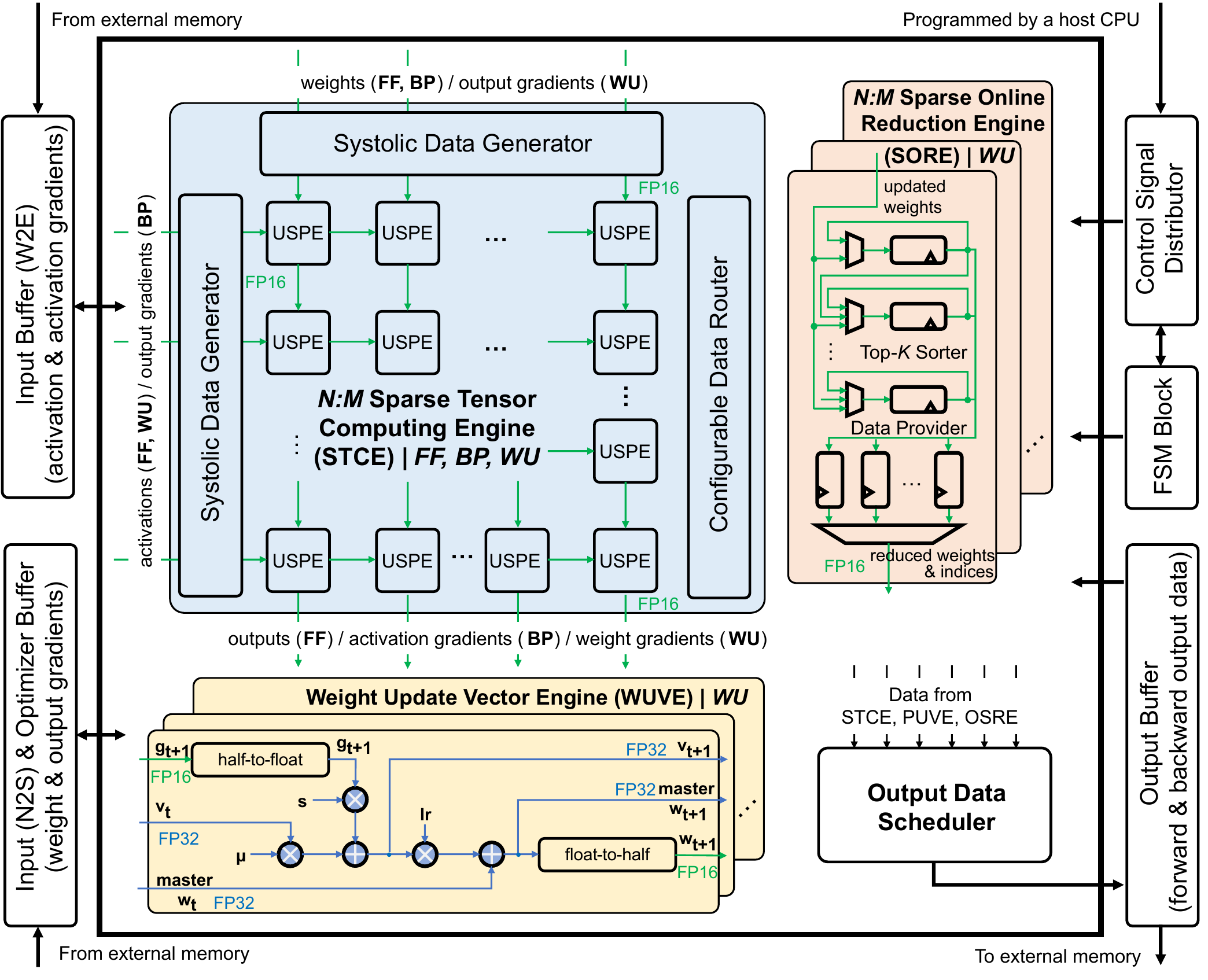}
	\caption{The overall microarchitecture of SAT is composed of three computing engines, namely STCE, WUVE, and SORE, respectively.} 
	\label{fig:arch}
\end{figure}

\subsection{Overall Architecture}

As shown in Fig.~\ref{fig:arch}, SAT consists of three major computing engines: 1) an \textit{N:M} sparse tensor computing engine (STCE), 2) a weight update vector engine (WUVE), and 3) a sparse online reduction engine (SORE).
STCE significantly boosts the computational efficiency of DNN training by efficiently unifying the MatMuls across FF, BP, and WU, and flexibly supporting both \textit{N:M} sparse and dense operations in its processing elements.
WUVE is a dedicated optimizer capable of updating weights through a mixed-precision scheme following NVIDIA Adaptive Mixed Precision (AMP) \cite{amp}, which can significantly reduce off-chip memory access.
SORE undertakes online \textit{N:M} sparse reduction operations by taking dense weights with a group size of \textit{M} as input and producing \textit{N:M} sparse weights along with corresponding indexes as output.
To improve the overall hardware performance, double-buffering is employed across all on-chip buffers to overlap the data transfer and computation.

\begin{figure} [htbp] 
	\centering
	\includegraphics[width=0.46\textwidth]{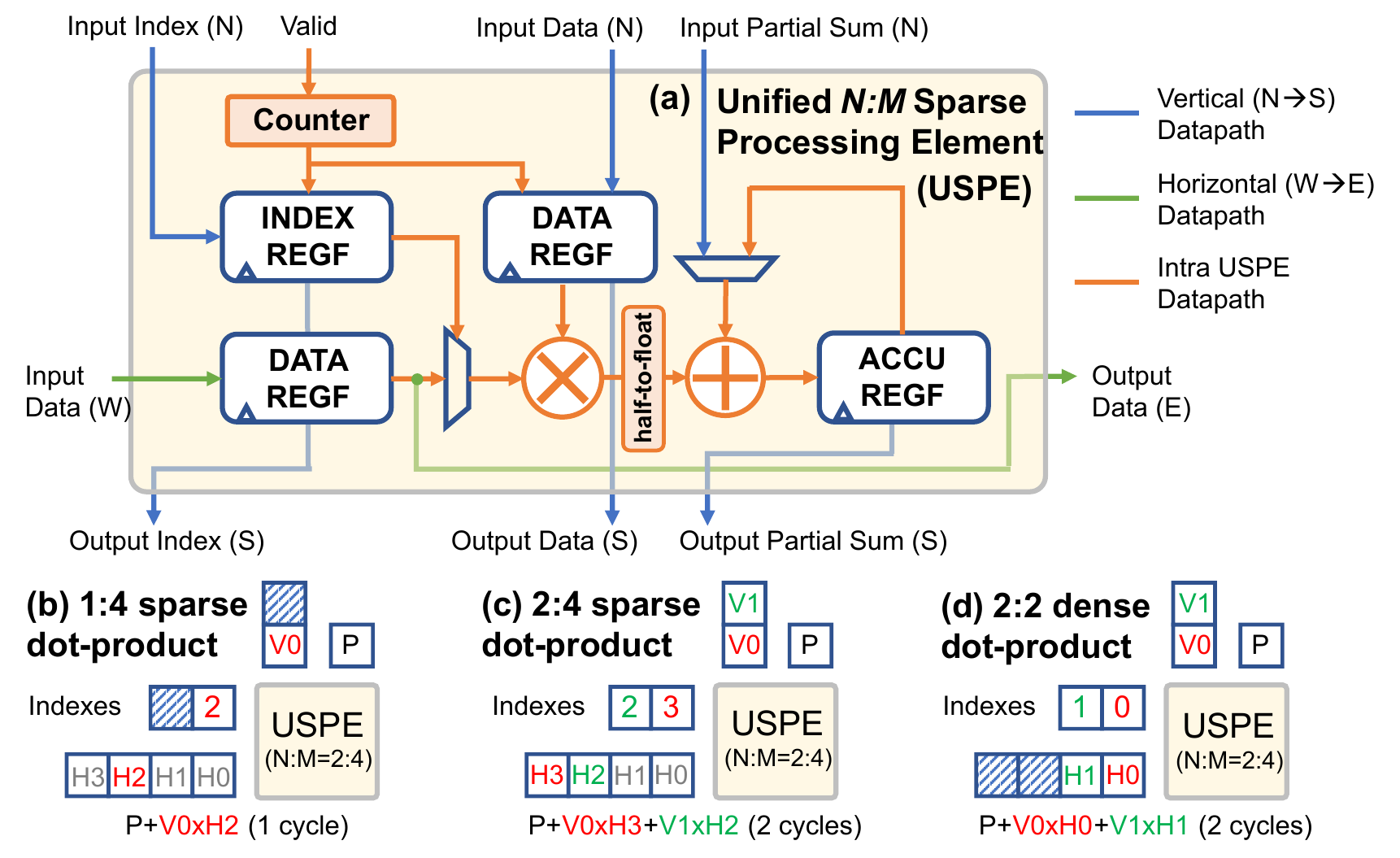}
	\caption{(a) The microarchitecture of unified \textit{N:M} sparse processing element (USPE). Example of 2:4 USPE performing (b) 1:4 sparse, (c) 2:4 sparse, and (d) 2:2 dense dot-product operations.} 
	\label{fig:arch_uspe}
\end{figure}

\subsection{Unified \textit{N:M} Sparse Processing Element}

STCE employs a systolic array composed of 32$\times$32 unified \textit{N:M} sparse processing elements (USPEs), which can be configured dynamically at runtime to perform \textit{N:M} sparse-dense or dense-dense products.
As depicted in Fig.~\ref{fig:arch_uspe}~(a), USPE comprises a task counter, an FP16 multiplier, an FP16-to-FP32 switcher, and an FP32 adder, in addition to four register files. 
These register files serve as temporary storage for input data received from the west and north, input valid indexes received from the north, and output accumulated partial results to the south.
During each cycle, the USPE is capable of multiplying two FP16 data and then adding its result to the input partial sum.
Both multiplier and adder in the USPE are pipelined by 3 stages to improve computational efficiency.

USPE is flexible enough to support diverse types of dot-product operations during DNN training.
Fig.~\ref{fig:arch_uspe}~(b)-(d) illustrate how the USPE performs dot-product operations across various \textit{N:M} sparse and dense configurations.
In order to accommodate varying \textit{N:M} sparsity, the USPE utilizes a value-serial computing approach, which facilitates the folding of the dot-product operation for an \textit{N:M} group into \textit{N} cycles. 
For instance, a 2:4 USPE can execute a 1:4 sparse dot-product within a cycle, and a 2:4 sparse dot-product within two cycles. 
Additionally, we decompose dense MatMul into multiple 2:2 dense dot-products, which are then assigned to USPEs. 
Each USPE can perform a 2:2 dense dot-product within two cycles.

\begin{figure} [htbp] 
	\centering
	\includegraphics[width=0.46\textwidth]{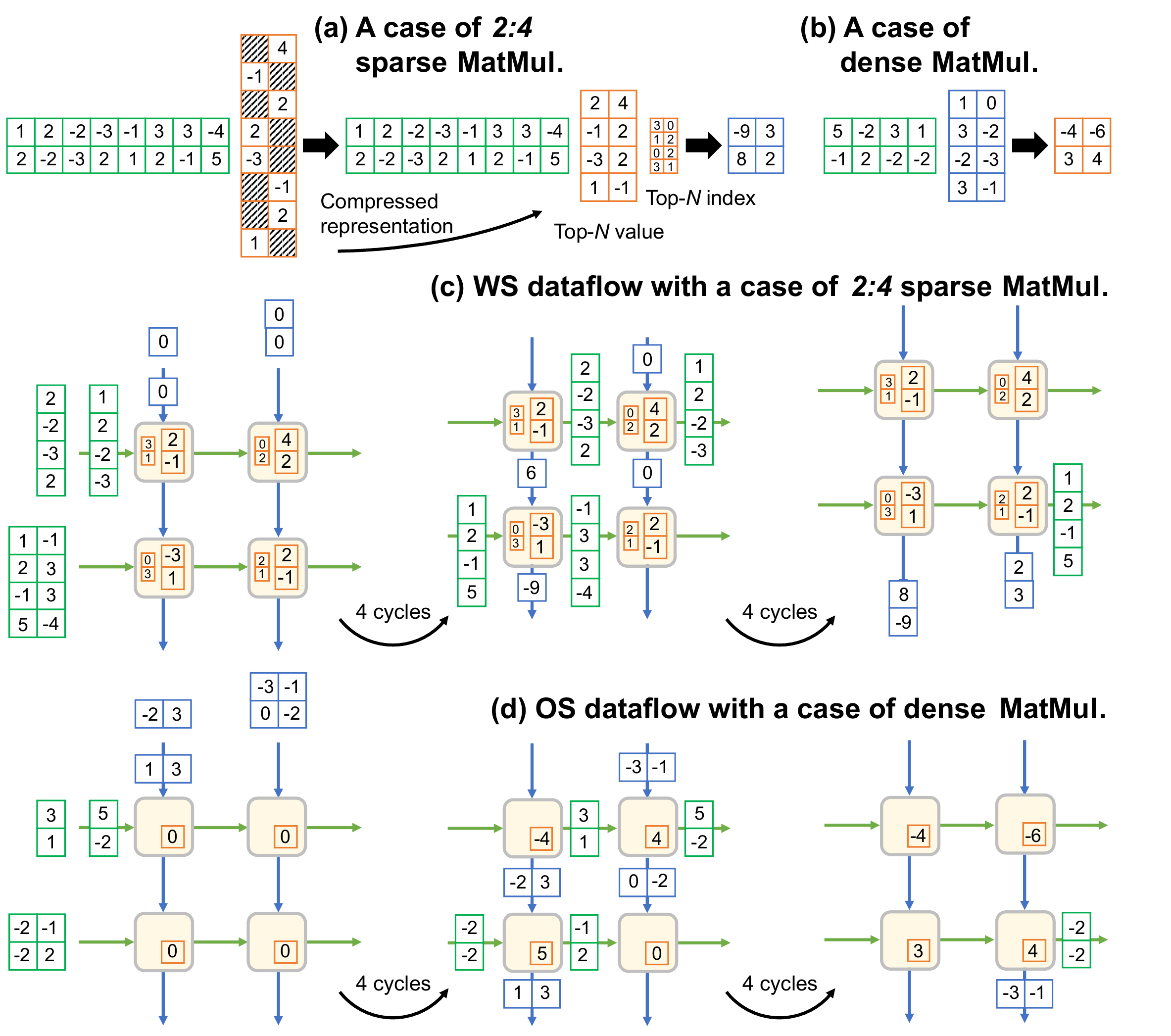}
	\caption{Example of STCE performing a 2:4 sparse MatMul in the WS dataflow and a dense MatMul in the OS dataflow.} 
	\label{fig:arch_systolic}
\end{figure}

\subsection{Flexible Systolic Interconnect}

\rOne{
STCE improves on previous works that employed weight-stationary (WS) \cite{jouppi2017datacenter, wu2022usystolic, kung2019maestro} or output-stationary (OS) \cite{venkataramanaiah2019automatic, venkataramanaiah2020fpga, fang2022algorithm} systolic architectures by leveraging a flexible systolic interconnect that is capable of dynamically switching between WS and OS dataflows on the fly, providing increased mapping space for MatMul operations and enabling efficient MatMuls across various computing patterns in FF, BP, and WU stages.
}

Fig.~\ref{fig:arch_systolic} presents how STCE equipped with the flexible systolic interconnect performs a 2:4 sparse MatMul in WS dataflow and a dense MatMul in OS dataflow.
In Fig.~\ref{fig:arch_systolic}~(a), we present a case of 2:4 sparse MatMul, where the sparse matrix is compactly packed by preserving only the two most significant values in each 2:4 group, along with their corresponding indexes.
When performing this operation in the WS dataflow, STCE first preloads the compact 2:4 weight groups to each USPE. 
The computation starts once the preload is complete.
To accomplish the dot-product operation of a 2:4 sparse group, each USPE in STCE consumes two cycles, after which it transfers its data from the west to the east and data from the north to the south. 
Fig.~\ref{fig:arch_systolic}~(c) illustrates the data transfer process every two times 2:4 sparse group computation tasks.
Due to the removal of pruned weight elements, STCE performs fewer operations in comparison with the dense task, thereby improving computational efficiency.
Fig.~\ref{fig:arch_systolic}~(b) is a case of dense MatMul.
When a dense MatMul is performed in OS dataflow, STCE with 2:4 USPEs streams the two input dense matrices from the west and north directions, respectively.
Every two cycles, a USPE performs 2:2 dense dot-product operations, and the computed data from the west is transferred to the east, while the data from the north is passed to the south.
The data transfer process is shown in Fig.~\ref{fig:arch_systolic}~(d), which illustrates the exchange of data every two times 2:2 dense group computation tasks.
The high utilization of USPEs during the computation stage enables STCE to achieve high computational efficiency for dense operations as well.
Finally, STCE sequentially pops its accumulation results to the south upon the computation is finished.

\subsection{\rTwo{Hardware Costs of STCE Enabling \textit{N:M} Sparse Operations}}
\rTwo{
To support \textit{N:M} sparse operations, STCE requires additional logic to support sparse decoding. 
It also requires more registers to support the storage overheads caused by the significant increase in input bandwidth in \textit{N:M} sparse MatMul compared to dense MatMul. 
As shown in Fig.~\ref{fig:arch_systolic}, for 2:4 STCE, each USPE requires 4 registers to store data from the west in the sparse mode, while in the dense mode, only two registers need to be enabled to complete the dense operations. 
In this case, the two disabled registers are the additional hardware overhead compared to a dense systolic array at the same scale. 
When enabling higher sparse ratios such as 2:8 and 2:16, the 2:4 STCE cannot directly implement 2:8 or 2:16 sparse operations, and needs to be reconfigured on FPGA. 
At higher \textit{N:M} sparse ratios, the register overhead per USPE in STCE will continue to grow, which may lead to disproportionate hardware costs, and the improvement in training accuracy may not be able to offset the continuous increase in hardware costs.
Therefore, the selection of \textit{N:M} sparsity is a trade-off between model accuracy and hardware cost.
}

\subsection{Weight Update Vector Engine}

\rTwo{As depicted in Fig.~\ref{fig:arch}, WUVE serves as a dedicated optimizer that employs momentum stochastic gradient descent (SGD) to update weights using the mixed precision scheme of NVIDIA AMP \cite{amp}.}
To update master parameters for the next training iteration, WUVE takes weight gradients in FP16 format and other master parameters in FP32 format as input.
It specifically elevates the numerical precision of weight gradients from FP16 to FP32 to minimize quantization errors in the WU step.
Moreover, WUVE provides 32 parallel lanes to improve computational efficiency, and each lane consists of three FP32 multipliers, two FP32 adders, one FP16-to-FP32 switcher, and one FP32-to-FP16 switcher.

\subsection{\textit{N:M} Sparse Online Reduction Engine}

\begin{figure} [htbp] 
	\centering
	\includegraphics[width=0.46\textwidth]{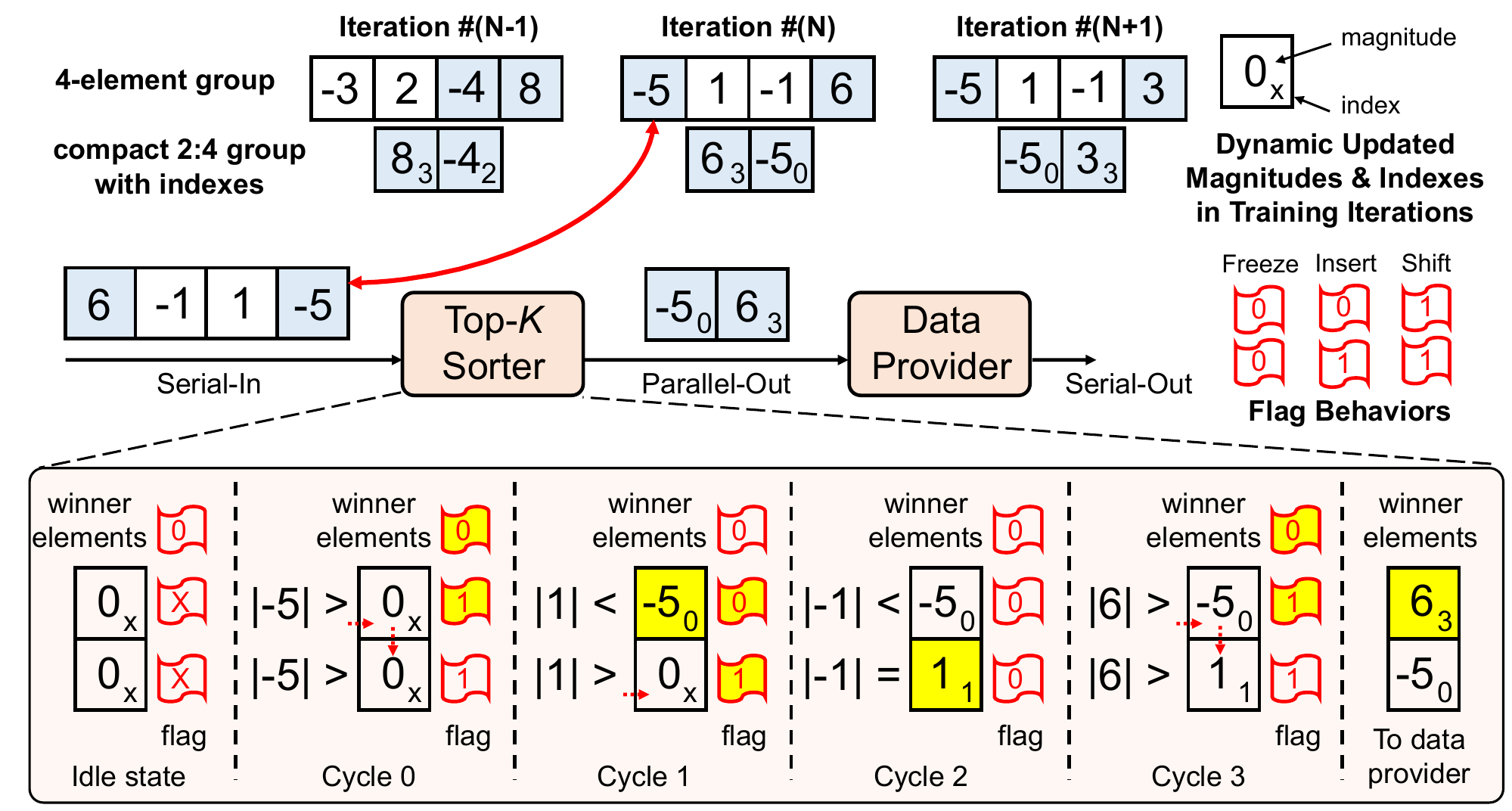}
	\caption{Example of the processing process of a 2:4 SORE. It generates 2:4 sparse groups with corresponding indexes in parallel using four cycles, starting with a four-element group as input.}
	\label{fig:arch_sore}
\end{figure}

\rTwo{Dedicated accelerators \cite{a100, fang2022algorithm}, designed for \textit{N:M} sparse inference acceleration, utilize \textit{N:M} sparse weights that are generated offline prior to inference.
However, during \textit{N:M} sparse training, weights, activations, and gradients are varied in every iteration, leading to dynamic updates of the preserved elements and their corresponding indexes in an \textit{N:M} element group, as illustrated in Fig.~\ref{fig:arch_sore}.
Therefore, to handle the dynamic updates of \textit{N:M} sparse elements during training, the dedicated module for \textit{N:M} sparse online reduction, namely SORE, is designed to enable the efficient generation of compact groups of \textit{N:M} sparse elements with their associated indexes.}

There are 32 parallel lanes in SORE, and each lane consists of a top-\textit{K} sorter and a data provider.
The top-\textit{K} sorter sequentially receives dense data in a group with a size of \textit{M}, and after \textit{M} cycles, the \textit{K} data sorted in the top-\textit{K} sorter with their indexes in the group are passed to the data provider.
The data provider, which is configurable to support all \textit{N} not larger than \textit{K}, sequentially outputs the top-\textit{K} elements in a group.
Fig.~\ref{fig:arch_sore} takes a 2:4 SORE as an example for illustration.
A 4-element group is sequentially provided to the top-\textit{K} sorter as input, and a 2:4 sparse group with corresponding indexes is generated in parallel to the data provider after 4 cycles.
\section{Dataflow Optimization} \label{sec:dataflow}

SAT is capable of effectively supporting \textit{N:M} sparse DNN training through its innovative design of STCE, which includes unified processing elements and a flexible interconnect, as well as SORE, which efficiently generates \textit{N:M} sparse data groups.
To further optimize the computational efficiency of SAT, we introduce several dataflow optimization methods to improve the utilization of STCE and the efficiency of SORE.

\subsection{Interleave Mapping of USPE}

\begin{figure} [htbp] 
	\centering
	\includegraphics[width=0.46\textwidth]{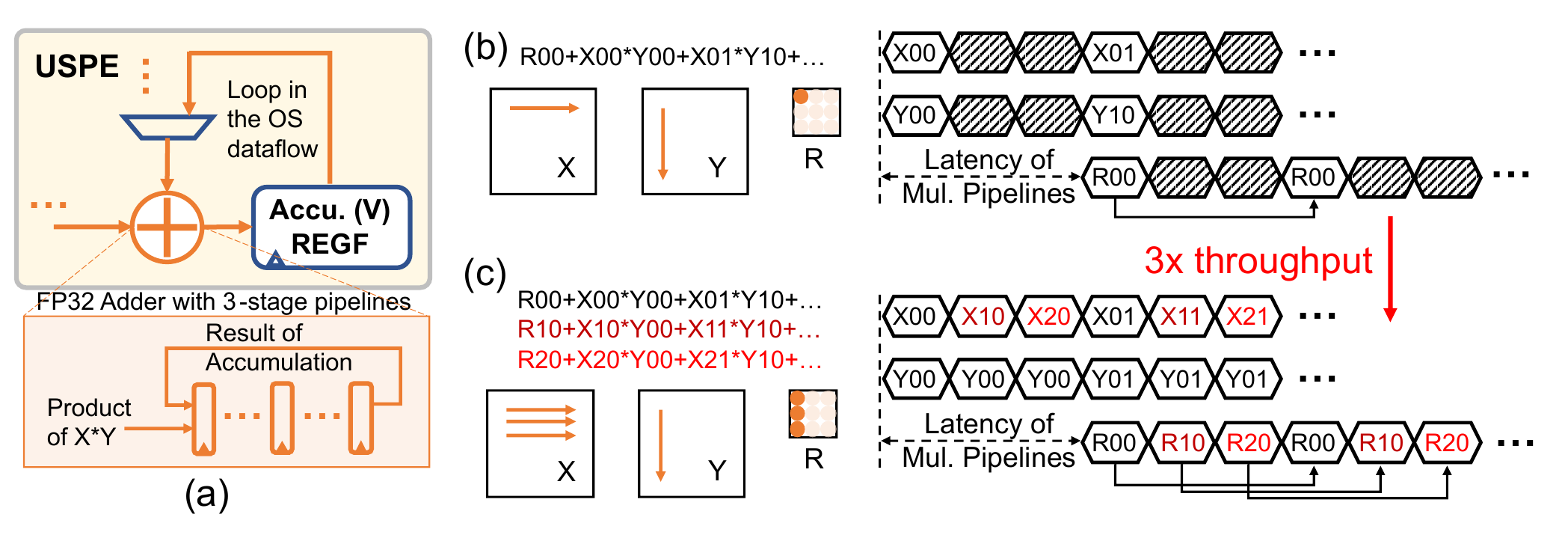}
	\caption{(a) The loop in USPE when employing OS dataflow slows down the computational efficiency of USPE using (b) the conventional systolic mapping strategy. By contrast, (c) the proposed interleave mapping strategy can improve 3x throughput by improving the utilization of USPE.} 
	\label{fig:df_interleave}
\end{figure}

To reduce the critical path of STCE and improve the operating frequency of SAT, both the multiplier and adder in USPE are deeply pipelined.
However, when STCE is configured as OS dataflow computing mode, there is an accumulation loop in USPE as shown in Fig.~\ref{fig:df_interleave}~(a).
This causes a computation stall when a dot-product operation is mapped to USPE, since the partial sums generated during dot-product operations are held in the pipelines until they reach the output stage. 
\rOne{As a result, the next dot-product operation has to wait until the previous one has cleared the pipeline, resulting in a three-cycle latency as shown in Fig.~\ref{fig:df_interleave}~(b).}

\rOne{By contrast, we propose an interleave mapping method for USPE that allows for the simultaneous processing of computationally independent operations during the accumulation loop. 
This helps to minimize the stall and improve the computational efficiency of USPE.}
As shown in Fig.~\ref{fig:df_interleave}~(c), three parallel dot-product operations are interleaved and fed into USPE, effectively filling up the pipelines.
By leveraging the proposed interleaving mapping method, USPE \rTwo{can} achieve $3\times$ throughput improvement when employing OS dataflow.

\subsection{Pre-generation of \textit{N:M} Sparse Elements}

\begin{figure} [htbp] 
	\centering
	\includegraphics[width=0.48\textwidth]{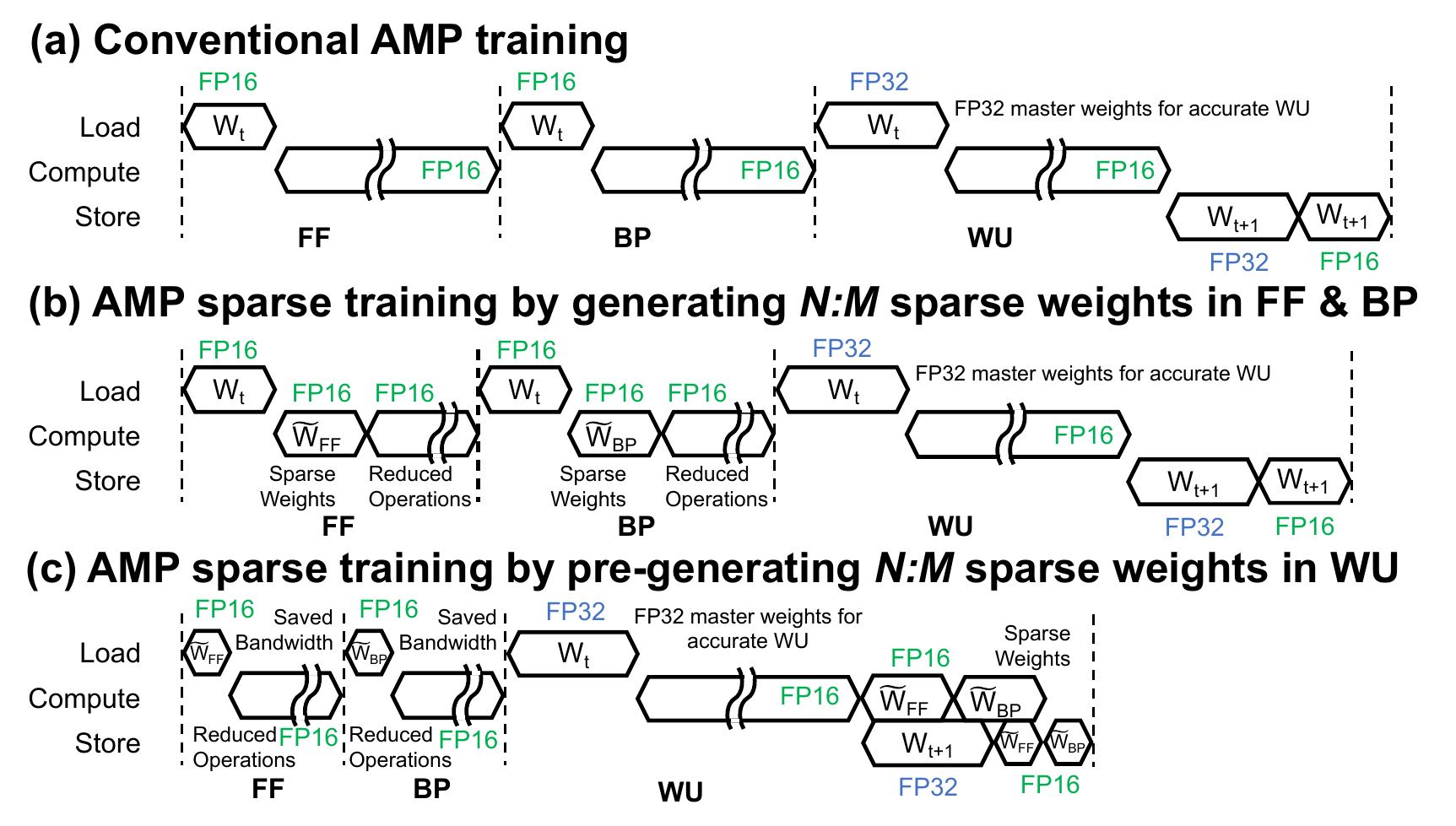}
	\caption{Weight dataflow schedule using (a) conventional AMP training, (b) sparse AMP training of BDWP with \textit{N:M} sparse generation in FF and BP, and (c) sparse AMP training of BDWP with \textit{N:M} sparse pre-generation in WU. Compared with (b), (c) saves external memory space, memory access bandwidth, and execution time at a high \textit{N:M} sparse ratio.}
	\label{fig:df_wgt_gen}
\end{figure}

\rOne{Pre-generation technique for \textit{N:M} sparse weights is proposed to boost the efficiency of \textit{N:M} sparse training with NVIDIA AMP \cite{amp}.}
AMP is a training scheme that optimizes the precision of the arithmetic operations used during training in DNNs.
Fig.~\ref{fig:df_wgt_gen}~(a) presents the computational steps to train a convolutional layer using AMP.
It can perform the forward pass with half-precision (FP16) arithmetic and accumulate and convert the gradients back to the original precision (FP32) before updating the weights.

\rOne{BDWP is taken as an example to integrate \textit{N:M} sparse training into AMP.}
\rOne{As shown in Fig.~\ref{fig:df_wgt_gen}~(b), \textit{N:M} sparse data can be generated after the weight tiles are loaded in FF and BP, and the computational cost is significantly reduced by skipping zero-value operations. 
However, this generation process can slow down computational efficiency, as the MatMul must wait for the generation of \textit{N:M} sparse weights.}
\rOne{Fig.~\ref{fig:df_wgt_gen}~(c) illustrates the working flow of our proposed pre-generation technique for \textit{N:M} sparse weights, which can improve the computational efficiency and storage requirement of \textit{N:M} sparse training.}
\rTwo{In WU stage, the FP32 weight updates are calculated in WUVE and then directly sent to SORE for \textit{N:M} sparse compression to obtain FP16 sparse weights. 
This process is fine-grained pipelined, thus achieving the overlap of computation and storage. In contrast, in Fig.~\ref{fig:df_wgt_gen}~(b), dense FP16 weights must be loaded from external memory and sent to SORE to obtain sparse weights before they can be sent to STCE for MatMul computation, which cannot achieve the overlap and affects the overall computation efficiency.}
In addition, compared to Fig.~\ref{fig:df_wgt_gen}~(b), Fig.~\ref{fig:df_wgt_gen}~(c) can save the bandwidth requirement by storing and loading the \textit{N:M} sparse weights.
\rOne{It requires to store for the \textit{N:M} weights $\widetilde{w}_{\text{FF}}$ and $\widetilde{w}_{\text{BP}}$ from its input and output channels.}
When its sparse ratio is higher than 50\%, the storage cost is also significantly lower than the conventional AMP training.

\begin{figure} [htbp] 
	\centering
	\includegraphics[width=0.48\textwidth]{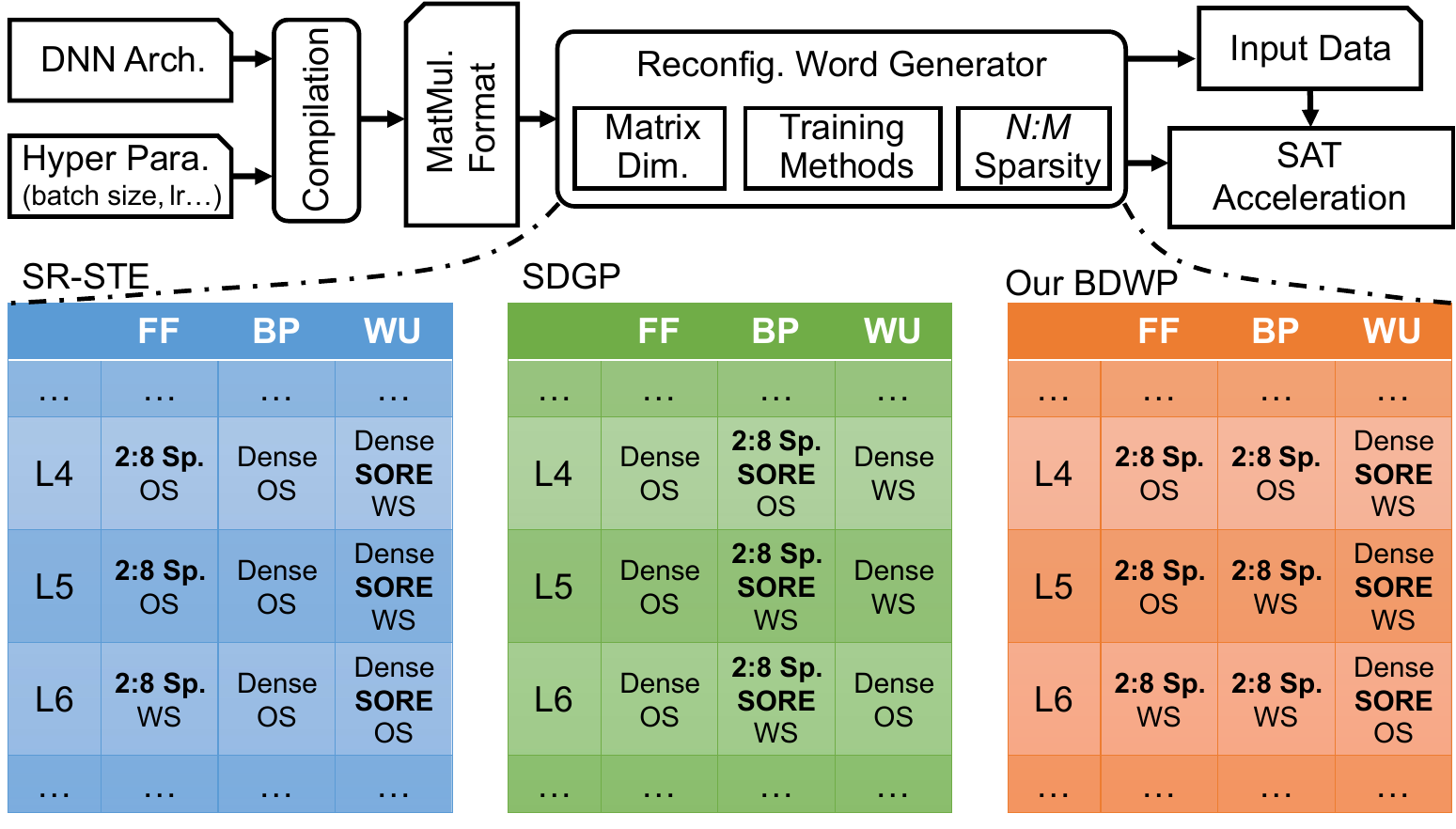}
	\caption{Enabling \textit{N:M} sparse DNN training on SAT with offline dataflow scheduling, which first transforms the DNN model into the MatMul format, and then generates per-layer configuration words for the three training stages.} 
	\label{fig:flex_scheduler}
    \vspace{-10pt}
\end{figure}

\subsection{Offline Dataflow Scheduling}

Fig.~\ref{fig:flex_scheduler} illustrates the implementation of \textit{N:M} sparse DNN training on SAT through flexible offline dataflow scheduling.
It involves transforming the DNN model into the MatMul format, and generating per-layer configuration words for the three training stages.
The reconfiguration word generator (RWG) takes the MatMul format of DNNs in three training stages as input and generates per-layer configuration words based on the selected \textit{N:M} sparse training method and the \textit{N:M} sparse ratio.
During training acceleration, SAT's controller fetches each layer's reconfiguration words gradually at the FF, BP, and WU stages and produces corresponding control signals for the other components.

\rTwo{RWG is the key component for improving the training throughput of SAT.}
\rTwo{Fig.~\ref{fig:flex_scheduler} presents how RWG produces reconfiguration words for ResNet18 when using three \textit{N:M} sparse training methods: \rTwo{SR-STE}, SDGP, and BDWP enabled with 2:8 sparse ratio.}
\rTwo{RWG first assigns \textit{N:M} sparse mode for FF, BP, and WU these three training stages based on the user-configured \textit{N:M} sparsity (e.g., 2:8) and the selected sparse training method. 
For example, for SR-STE, RWG will determine the FF stage of the network layer as 2:8 sparse training, and the BP and WU stages as dense training. For BDWP, which introduces \textit{N:M} sparsity in both directions, RWG will determine the FF and BP stages as 2:8 sparse training, and WU as dense training.}
\rTwo{Next, RWG allocates the method of generating \textit{N:M} sparse data. 
The pre-generation of \textit{N:M} sparse elements in the WU stage is prioritized due to its significant advantages.
If pre-generated features are not available, the corresponding layers need to generate \textit{N:M} sparse elements during the FF and BP computing stages. 
For example, SR-STE and BDWP, which allow the pre-generation of sparse weights, prompt the RWG to schedule SORE within the WU stage. Conversely, SDGP, which prunes input gradients during the BP stage, requires RWG to schedule SORE within the BP stage.}
\rTwo{Finally, RWG predicts the computational utilization of SAT based on the scale of transformed MatMul of each network layer in advance, so that it can arrange the superior dataflow and the output data layout rules for each layer. As shown in Fig.~\ref{fig:flex_scheduler}, for BDWP, within the 4th layer, RWG calculates the hardware utilization of OS and WS in the FF, BP, and WU phases, and based on predicted results, the OS, OS, and WS dataflows are allocated to the three phases, respectively.}
\rTwo{By assigning the above three phases, RWG effectively improves the computational utilization of each network layer, thereby significantly improving the training throughput of SAT.}

\begin{table}[htbp]
\centering
\caption{\rTwo{From-Scratch Training Setup of Evaluated DNNs}}
\label{tab:training_setup}
\resizebox{0.48\textwidth}{!}{%
\begin{threeparttable}
\begin{tabular}{@{}ccccccc@{}}
\toprule
\rTwo{Model}   & \rTwo{Dataset}       & \rTwo{Optimizer}    & \rTwo{EP\tnote{$\dagger$}} & \rTwo{BS\tnote{$\dagger$}}  & \rTwo{LR\tnote{$\dagger$}}   & \rTwo{WD\tnote{$\dagger$}}   \\ \midrule
\rTwo{ResNet9}     & \rTwo{CIFAR-10}      & \rTwo{Momentum SGD} & \rTwo{150}    & \rTwo{512} & \rTwo{0.5}  & \rTwo{5e-4} \\
\rTwo{ViT}         & \rTwo{CIFAR-100}     & \rTwo{Momentum SGD} & \rTwo{150}    & \rTwo{512} & \rTwo{0.1}  & \rTwo{5e-4} \\
\rTwo{VGG19}       & \rTwo{CIFAR-100}     & \rTwo{Momentum SGD} & \rTwo{150}    & \rTwo{512} & \rTwo{0.1}  & \rTwo{5e-4} \\
\rTwo{ResNet18}    & \rTwo{Tiny ImageNet} & \rTwo{Momentum SGD} & \rTwo{88}     & \rTwo{512} & \rTwo{0.05} & \rTwo{5e-3} \\
\rTwo{ResNet50}    & \rTwo{ImageNet}      & \rTwo{Momentum SGD} & \rTwo{120}    & \rTwo{256} & \rTwo{0.1}  & \rTwo{5e-5} \\ \bottomrule
\end{tabular}%
\begin{tablenotes}
    \item[$\dagger$] \rTwo{EP: epochs; BS: batch size; LR: initial learning rate; WD: weight decay.}
\end{tablenotes}
\end{threeparttable}
}
\end{table}

\section{Experimental Results} \label{sec:res}

\subsection{Experimental Setup}

\textbf{Benchmarks.}
\rTwo{To evaluate the algorithmic and hardware performance of our proposed \textit{N:M} sparse training scheme, we choose five typical DNN models on four popular datasets for from-scratch training. The details are shown in Table~\ref{tab:training_setup}.}
\rTwo{The ResNet9, ResNet18, ResNet50 \cite{he2016deep} and VGG19 \cite{simonyan2014very} are the conventional convolutional neural networks (CNNs), and the vision Transformer (ViT) \cite{dosovitskiy2021an} is a novel DNN architecture utilizing the attention mechanism.}

\textbf{Software Implementation.}
\rTwo{We train the DNN benchmarks with the hyper-parameter settings shown in Table~\ref{tab:training_setup} using PyTorch v1.10 with mixed-precision training support~\cite{amp}.}
Note that, in each training iteration, \textit{N:M} sparsity is applied to all convolutional layers except for the first layer of \rTwo{evaluated CNNs}, and all linear layers in the Transformer blocks of ViT.
\rTwo{Excluding the first convolutional layer from \textit{N:M} sparsity aligns with SDGP's experimental setup \cite{mcdanel2022accelerating} and quantization-related practices~\cite{choukroun2019low} due to the first layer's sensitivity to accuracy impacts arising from its limited input channels.}
\textit{N:M} sparse patterns are introduced since the first training step and kept updated for every iteration until the end of training.

\textbf{Hardware Implementation.}
SAT is implemented in SystemVerilog with the help of hardware components from the PULP platform~\cite{rossi2015pulp} and the BaseJump standard template library~\cite{taylor2018basejump}. 
We evaluate hardware performance using a Xilinx Virtex UltraScale+ VCU1525 card with an XCVU9P FPGA, with Xilinx Vivado 2018.2 at a clock frequency of 200 MHz. 
We measure power consumption using the Xilinx Power Estimator tool.
\rOne{
Moreover, we validate speed performance using a cycle-accurate performance model that is cross-validated with RTL simulation results following methods in \cite{fan2022adaptable, venkataramanaiah2020fpga}.
The memory accesses to external memory are also considered.
}

\begin{table*}[htbp]
\centering
\caption{\rTwo{Accuracy Comparison Using Various \textit{N:M} Sparse Training Schemes}}
\label{tab:algo_res}
\resizebox{\textwidth}{!}{%
\begin{tabular}{@{}ccc|ccc|ccc|ccc|ccc|ccc@{}}
\toprule
\textbf{}                      &               &                                                                         & \multicolumn{3}{c|}{\textbf{ResNet9 on CIFAR-10}}                                                                                                                                                                                               & \multicolumn{3}{c|}{\rTwo{\textbf{VGG19 on CIFAR-100}}}                                                                                                                                                                                                & \multicolumn{3}{c|}{\textbf{ViT on CIFAR-100}}                                                                                                                                                                                                  & \multicolumn{3}{c|}{\textbf{ResNet18 on Tiny ImageNet}}                                                                                                                                                                                         & \multicolumn{3}{c}{\rTwo{\textbf{ResNet50 on ImageNet}}}                                                                                                                                                                                               \\ \midrule
\textbf{Method}                & \textbf{Pat.} & \textbf{\begin{tabular}[c]{@{}c@{}}Sparsity\\ in \\ FW\&BW\end{tabular}} & \textbf{\begin{tabular}[c]{@{}c@{}}Train.\\ FLOPS\\ ($\times10^{16}$)\end{tabular}} & \textbf{\begin{tabular}[c]{@{}c@{}}Infer.\\ FLOPS\\ ($\times10^{9}$)\end{tabular}} & \textbf{\begin{tabular}[c]{@{}c@{}}Top-1\\ Acc.\\ (\%)\end{tabular}} & \textbf{\begin{tabular}[c]{@{}c@{}}Train.\\ FLOPS\\ ($\times10^{15}$)\end{tabular}} & \textbf{\begin{tabular}[c]{@{}c@{}}Infer.\\ FLOPS\\ ($\times10^{8}$)\end{tabular}} & \textbf{\begin{tabular}[c]{@{}c@{}}Top-1\\ Acc.\\ (\%)\end{tabular}} & \textbf{\begin{tabular}[c]{@{}c@{}}Train.\\ FLOPS\\ ($\times10^{16}$)\end{tabular}} & \textbf{\begin{tabular}[c]{@{}c@{}}Infer.\\ FLOPS\\ ($\times10^{8}$)\end{tabular}} & \textbf{\begin{tabular}[c]{@{}c@{}}Top-1\\ Acc.\\ (\%)\end{tabular}} & \textbf{\begin{tabular}[c]{@{}c@{}}Train.\\ FLOPS\\ ($\times10^{16}$)\end{tabular}} & \textbf{\begin{tabular}[c]{@{}c@{}}Infer.\\ FLOPS\\ ($\times10^{9}$)\end{tabular}} & \textbf{\begin{tabular}[c]{@{}c@{}}Top-1\\ Acc.\\ (\%)\end{tabular}} & \textbf{\begin{tabular}[c]{@{}c@{}}Train.\\ FLOPS\\ ($\times10^{18}$)\end{tabular}} & \textbf{\begin{tabular}[c]{@{}c@{}}Infer.\\ FLOPS\\ ($\times10^{9}$)\end{tabular}} & \textbf{\begin{tabular}[c]{@{}c@{}}Top-1\\ Acc.\\ (\%)\end{tabular}} \\ \midrule
Baseline                       & -         & \ding{56}~~~~\ding{56}                                                      & 2.62                                                                                & 1.16                                                                               & 95.27                                                                & 9.00                                                                                & 4.00                                                                               & 72.23                                                                & 1.45                                                                                & 6.43                                                                               & 60.81                                                                & 4.82                                                                                & 1.83                                                                               & 65.46                                                                & 1.91                                                                                & 4.14                                                                               & 76.72                                                                \\
\cite{zhou2021learning}        & 2:4           & \ding{52}~~~~\ding{56}                                                      & 2.19                                                                                & 0.59                                                                               & 95.18                                                                & 7.52                                                                                & 2.02                                                                               & 73.04                                                                & 1.22                                                                                & 3.36                                                                               & 60.37                                                                & 4.07                                                                                & 0.98                                                                               & 65.43                                                                & 1.61                                                                                & 2.16                                                                               & 76.52                                                                \\
\cite{mcdanel2022accelerating} & 2:4           & \ding{56}~~~~\ding{52}                                                      & 2.19                                                                                & 1.16                                                                               & 95.11                                                                & 7.52                                                                                & 4.00                                                                               & 72.01                                                                & 1.22                                                                                & 6.43                                                                               & 57.67                                                                & 4.07                                                                                & 1.83                                                                               & 64.99                                                                & 1.61                                                                                & 4.14                                                                               & N/A                                                            \\
\textbf{BDWP}                  & \textbf{2:4}  & \textbf{\ding{52}~~~~\ding{52}}                                             & \textbf{1.75}                                                                       & \textbf{0.59}                                                                      & \textbf{95.10}                                                       & \textbf{6.03}                                                                       & \textbf{2.02}                                                                      & \textbf{72.21}                                                       & \textbf{0.99}                                                                       & \textbf{3.36}                                                                      & \textbf{60.85}                                                       & \textbf{3.33}                                                                       & \textbf{0.98}                                                                      & \textbf{65.40}                                                       & \textbf{1.30}                                                                       & \textbf{2.16}                                                                      & \textbf{76.80}                                                       \\
\cite{zhou2021learning}        & 2:8           & \ding{52}~~~~\ding{56}                                                      & 1.97                                                                                & 0.30                                                                               & 95.18                                                                & 6.78                                                                                & 1.03                                                                               & 72.78                                                                & 1.10                                                                                & 1.83                                                                               & 59.55                                                                & 3.70                                                                                & 0.55                                                                               & 65.04                                                                & 1.45                                                                                & 1.17                                                                               & 75.88                                                                \\
\cite{mcdanel2022accelerating} & 2:8           & \ding{56}~~~~\ding{52}                                                      & 1.97                                                                                & 1.16                                                                               & 95.18                                                                & 6.78                                                                                & 4.00                                                                               & 71.25                                                                & 1.10                                                                                & 6.43                                                                               & 46.10                                                                & 3.70                                                                                & 1.83                                                                               & 62.40                                                                & 1.45                                                                                & 4.14                                                                               & N/A                                                            \\
\textbf{BDWP}                  & \textbf{2:8}  & \textbf{\ding{52}~~~~\ding{52}}                                             & \textbf{1.32}                                                                       & \textbf{0.30}                                                                      & \textbf{95.18}                                                       & \textbf{4.55}                                                                       & \textbf{1.03}                                                                      & \textbf{72.32}                                                       & \textbf{0.76}                                                                       & \textbf{1.83}                                                                      & \textbf{59.60}                                                       & \textbf{2.58}                                                                       & \textbf{0.55}                                                                      & \textbf{65.14}                                                       & \textbf{1.00}                                                                       & \textbf{1.17}                                                                      & \textbf{75.44}                                                       \\
\cite{zhou2021learning}        & 2:16          & \ding{52}~~~~\ding{56}                                                      & 1.86                                                                                & 0.15                                                                               & 95.15                                                                & 6.40                                                                                & 0.53                                                                               & 72.18                                                                & 1.04                                                                                & 1.06                                                                               & 55.72                                                                & 3.52                                                                                & 0.34                                                                               & 63.75                                                                & 1.38                                                                                & 0.67                                                                               & 74.75                                                                \\
\cite{mcdanel2022accelerating} & 2:16          & \ding{56}~~~~\ding{52}                                                      & 1.86                                                                                & 1.16                                                                               & 94.91                                                                & 6.40                                                                                & 4.00                                                                               & 69.95                                                                & 1.04                                                                                & 6.43                                                                               & 38.37                                                                & 3.52                                                                                & 1.83                                                                               & 48.00                                                                & 1.38                                                                                & 4.14                                                                               & N/A                                                            \\
\textbf{BDWP}                  & \textbf{2:16} & \textbf{\ding{52}~~~~\ding{52}}                                             & \textbf{1.10}                                                                       & \textbf{0.15}                                                                      & \textbf{95.16}                                                       & \textbf{3.80}                                                                       & \textbf{0.53}                                                                      & \textbf{72.04}                                                       & \textbf{0.64}                                                                       & \textbf{1.06}                                                                      & \textbf{55.70}                                                       & \textbf{2.21}                                                                       & \textbf{0.34}                                                                      & \textbf{63.94}                                                       & \textbf{0.84}                                                                       & \textbf{0.67}                                                                      & \textbf{74.24}                                                       \\ \bottomrule
\end{tabular}%
}
\end{table*}

\subsection{Algorithmic Performance}

\rTwo{We compare the Top-1 accuracy of BDWP against the other state-of-the-art \textit{N:M} sparse training methods, including SDGP \cite{mcdanel2022accelerating} and \rTwo{SR-STE} \cite{zhou2021learning}.}
To evaluate the robustness of these sparse training methods, we inherit the hyperparameter settings of the baseline to BDWP, SDGP, and \rTwo{SR-STE}.
\rTwo{Notably, all experiments are repeated three times for reliability, except for ResNet50 on ImageNet, which has a single run due to limited available computational resources.}

\rTwo{Table~\ref{tab:algo_res} shows how the different \textit{N:M} sparse training methods affect the convergent model accuracy.}
'Pat.' is short for \textit{N:M} sparsity pattern, while 'FW' and 'BW' refer to introduced \textit{N:M} sparse patterns from the forward and backward passes, respectively.
In most cases, BDWP results in the best performance with the lowest training FLOPS compared to SDGP and \rTwo{SR-STE}, indicating BDWP is an effective method to reduce computational costs without sacrificing accuracy.
\rTwo{As shown in Table~\ref{tab:algo_res}, with a 2:8 sparse ratio, BDWP achieves an average theoretical computational reduction of 1.93$\times$ across evaluated training tasks compared to the dense training scheme.
The significant reduction of computational operations is coupled with negligible impact on model convergence accuracy, showing an average loss of only 0.56\%.
Moreover, the required number of operations for inference significantly reduces by 3.54$\times$ on average.}

\begin{figure} [htbp] 
	\centering
	\includegraphics[width=0.46\textwidth]{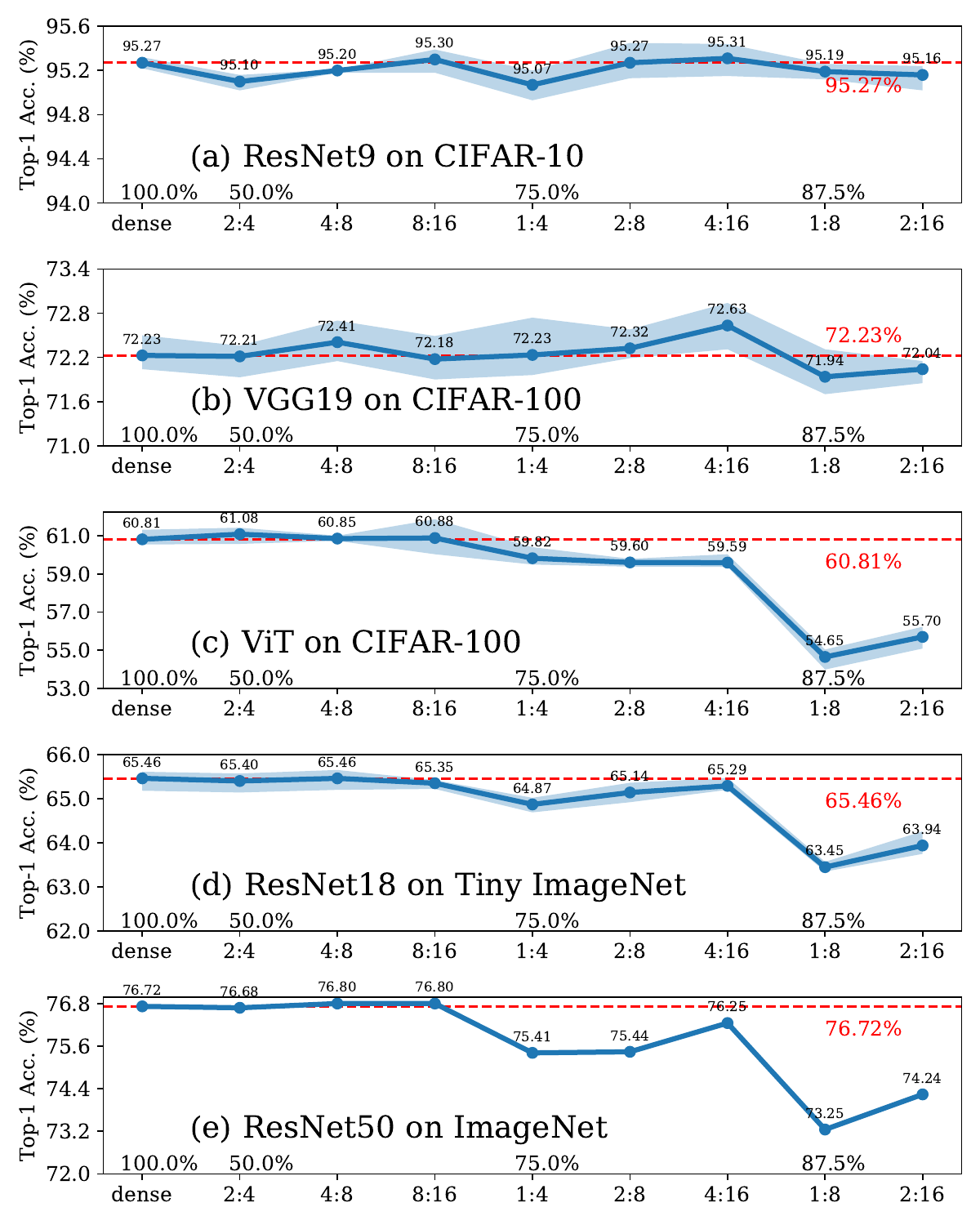}
	\caption{\rTwo{Impact of various \textit{N:M} sparse ratios on model convergent accuracy when employing BDWP sparse training.}} 
	\label{fig:res_nm_impact}
\end{figure}

\rTwo{Fig.~\ref{fig:res_nm_impact} illustrates the impact of various \textit{N:M} sparse ratios on model convergent accuracy when employing BDWP sparse training.
Experimental results show that the larger the \textit{N:M} sparse ratio, the less likely the model is to overfit the dataset. For example, ResNet9 on CIFAR-10 is less prone to overfitting, so it can tolerate a higher \textit{N:M} sparse ratio without losing too much accuracy. However, for models that are less prone to overfitting on selected datasets, such as ViT, ResNet18, and ResNet50, a higher \textit{N:M} sparse ratio, up to 87.5\%, may result in a slight decrease in accuracy. This is because these models are less tolerant of the loss of representation capability due to sparsity.
Additionally, the impact of \textit{M} on model accuracy varies depending on the level of sparsity. 
For instance, at lower sparsity levels, like 50\%, the choice of \textit{M} may not have a significant impact on accuracy at the same \textit{N:M} sparsity ratio. 
On the other hand, for higher sparsity ratios, up to 87.5\%, a larger \textit{M} can provide more flexible pattern choices, leading to better accuracy performance. 
For example, when comparing the ViT, ResNet18, and ResNet50 models, it has been observed that the convergence accuracy of a 2:16 sparse model with a sparsity ratio of 87.5\% is higher than that of a 1:8 sparse model.
Therefore, the choice of \textit{M} should be carefully considered when designing \textit{N:M} sparse models to achieve optimal accuracy and sparsity trade-offs.}

\subsection{Hardware Resource Consumption}
\begin{figure} [tbp] 
    \centering
    \includegraphics[width=0.48\textwidth]{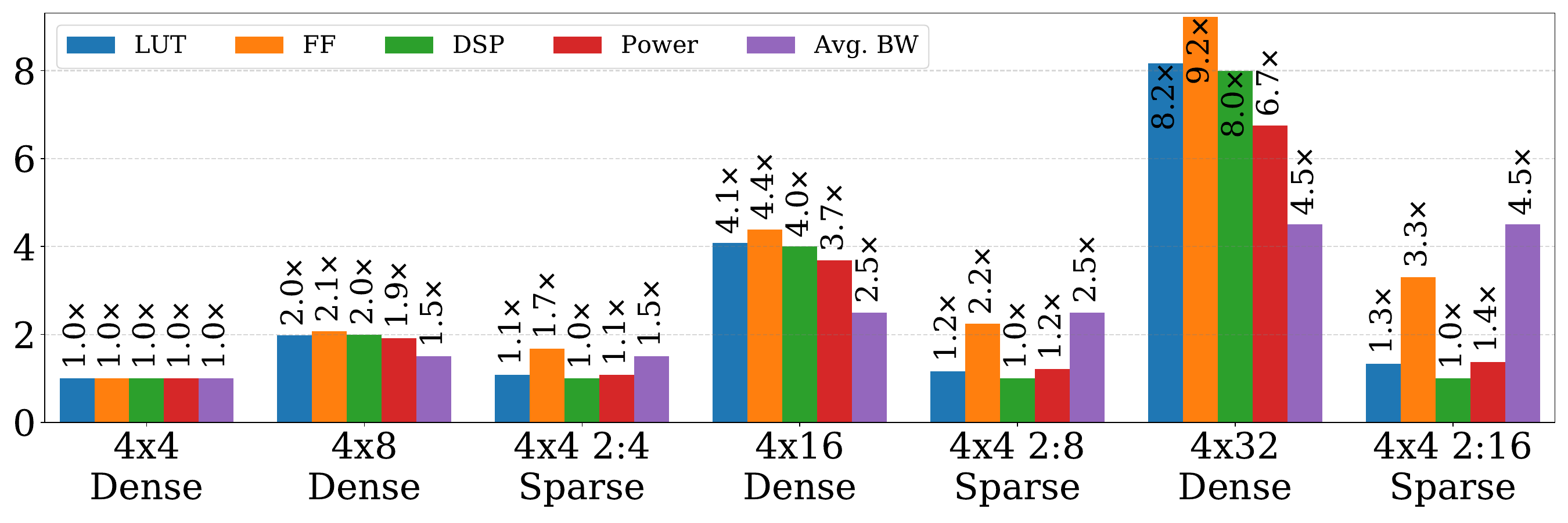}
    \caption{\rTwo{Hardware resource comparison between multiple dense systolic arrays and STCE of various \textit{N:M} sparse ratios.}}
    \label{fig:res_STCE}
\end{figure}

\rTwo{To precisely understand the hardware overhead of STCE, we conduct an experiment using a 4$\times$4 dense systolic array as a baseline, along with 4$\times$4 STCEs under various \textit{N:M} sparse configurations.
Additionally, for a fair comparison, we implement the other baseline systolic arrays with the same throughput of \textit{N:M} STCEs.
To support sparse \textit{N:M} operations, STCE requires additional LUT overhead for supporting sparse indexes, additional FF overhead for storing \textit{N:M} data groups, and corresponding power overhead compared to the dense baseline.
Experimental results are shown in Fig.~\ref{fig:res_STCE}. 
Compared to the 4$\times$4 dense baseline, 2:4, 2:8, and 2:16 STCEs increase the LUT overhead by 1.1$\times$, 1.2$\times$, and 1.3$\times$, respectively, while the FF overhead increases significantly by 1.7$\times$, 2.2$\times$, and 3.3$\times$.
However, these additional hardware costs are highly worthwhile when comparing STCE with the dense systolic arrays operating at the same throughput scale.
As shown in Fig.~\ref{fig:res_STCE}, 2:8 STCE has significantly lower hardware overheads than 4$\times$16 dense systolic array, with 3.4$\times$ lower LUT, 2.0$\times$ lower FF, 4.0$\times$ lower DSP, and 3.1$\times$ lower power consumptions, which shows that STCE is a promising architecture for performing sparse \textit{N:M} operations.}

\begin{table}[htbp]
\centering
\caption{SAT Resource Breakdown on Xilinx VCU1525}
\label{tab:sat_breakdown}
\resizebox{0.44\textwidth}{!}{%
\begin{tabular}{@{}c|cccc@{}}
\toprule
Component           & Logic                                                 & Registers                                             & \begin{tabular}[c]{@{}c@{}}Memory\\ blocks\end{tabular} & \begin{tabular}[c]{@{}c@{}}DSP\\ blocks\end{tabular}  \\ \midrule
STCE                & 389K                                                  & 589K                                                  & 0                                                       & 1024                                                  \\
WUVE                & 40K                                                   & 20K                                                   & 0                                                       & 192                                                   \\
SORE                & 3K                                                    & 5K                                                    & 0                                                       & 0                                                     \\
Input Buffer (W2E)  & 0                                                     & 0                                                     & 128                                                     & 0                                                     \\
Input Buffer (N2S)  & 0                                                     & 0                                                     & 38                                                      & 0                                                     \\
Output Buffer (N2S) & 0                                                     & 0                                                     & 38                                                      & 0                                                     \\
Optimizer Buffer    & 0                                                     & 0                                                     & 64                                                      & 0                                                     \\
Others              & 257K                                                  & 358K                                                  & 443                                                     & 12                                                    \\ \midrule
Total               & \begin{tabular}[c]{@{}c@{}}689K\\ (58\%)\end{tabular} & \begin{tabular}[c]{@{}c@{}}972K\\ (41\%)\end{tabular} & \begin{tabular}[c]{@{}c@{}}711\\ (23\%)\end{tabular}    & \begin{tabular}[c]{@{}c@{}}1228\\ (18\%)\end{tabular} \\ \bottomrule
\end{tabular}%
}
\end{table}

\rTwo{Based on the comprehensive trade-off from the algorithm and hardware perspectives, we select a 2:8 sparsity pattern in the following hardware implementation of SAT.} Table~\ref{tab:sat_breakdown} presents the resource consumption, where the 'others' row includes DDR4 controller, PCIe DMA, the interconnect from DDR to SAT, and other auxiliary components. 
Due to the adoption of 2:8 sparse patterns, the number of memory banks for W2E buffer needs to be expanded to four times that of N2S buffer, resulting in the use of 128 banks. 
Additionally, N2S buffer requires additional storage space to store sparse indexes, and therefore a total of 38 banks are used for both N2S input and output buffers. 
Furthermore, the optimizer buffer needs to store weight update parameters for 64 banks.

As shown in Table~\ref{tab:sat_breakdown}, STCE dominates DSP consumption of SAT since it is the computing core for computational intensive MatMuls that transformed from convolutional or linear layers. 
STCE also takes the majority of the register consumption since there are multiple pipelines in USPEs to shorten the critical path and improve computational throughput. 
Furthermore, SORE, the sparse online reduction engine, is an area-efficient hardware solution for enabling \textit{N:M} sparse online reduction capability in SAT, as it consumes less than 1\% of the resources utilized by STCE.

\begin{figure*} [htbp] 
	\centering
	\includegraphics[width=\textwidth]{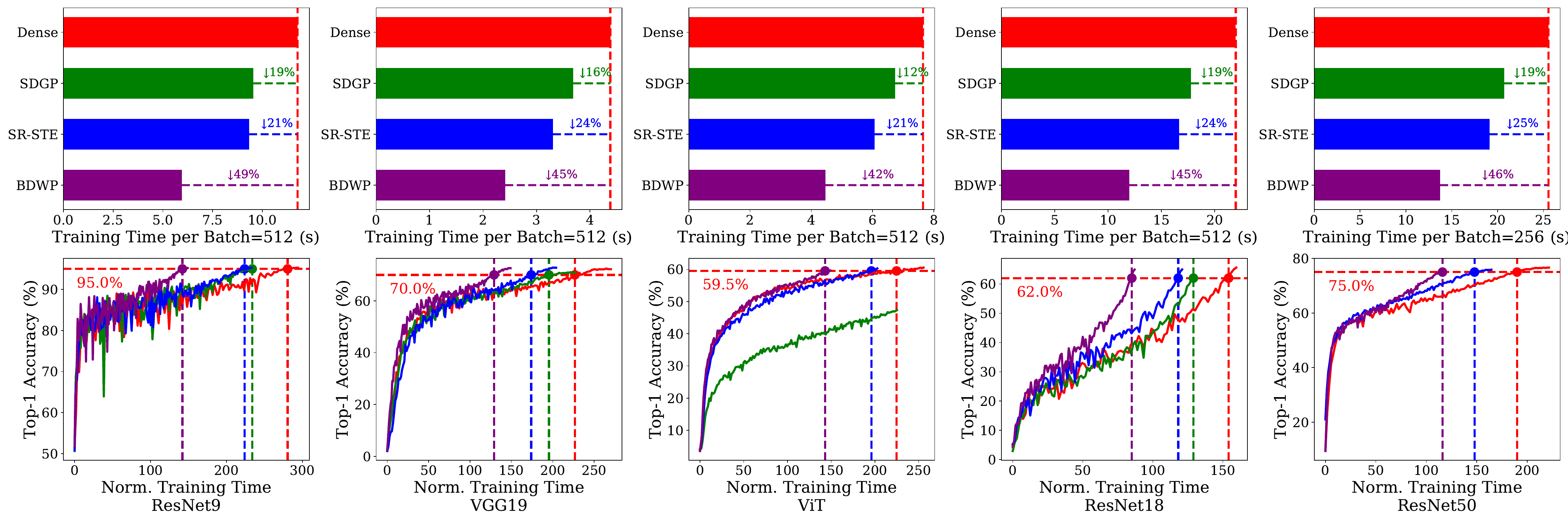}
    \caption{\rTwo{Required training time of \rTwo{SR-STE}, SDGP, and our BDWP on SAT for ResNet9, ViT, VGG19, ResNet18, and ResNet50, respectively.}}
	\label{fig:res_baseline}
\end{figure*}

\subsection{Training Efficiency}

To evaluate the training efficiency of the proposed training scheme composed of BDWP and SAT, Time-To-Accuracy (TTA) metric \cite{mcdanel2022accelerating, zhang2022fast} is used in comparison with other training methods, including conventional dense training, SDGP, and \rTwo{SR-STE}. 
SAT enables 2:8 sparse acceleration for SDGP, \rTwo{SR-STE}, and BDWP.
\rTwo{As depicted in the upper part of Fig.~\ref{fig:res_baseline}, SAT with 2:8 BDWP training achieves an average of 46\% reduction in single-batch training times compared to dense training, which corresponds to a significant 1.82$\times$ speedup per batch.}
\rTwo{Furthermore, the introduction of sparsity during training can impact the speed of model convergence, affecting the overall acceleration of sparse training. To ensure a fair comparison, the lower part of Fig.~\ref{fig:res_baseline} shows the model convergence curves during 2:8 BDWP sparse training compared to dense training on SAT, where training time is normalized by the required time for a training epoch of BDWP.
It reveals an average practical speedup of 1.75$\times$, which highlights the combined contribution of the sparse training algorithm and the hardware accelerator, with the algorithm reducing the computation and the hardware accelerating the process.}

\begin{figure} [htbp] 
	\centering
	\includegraphics[width=0.5\textwidth]{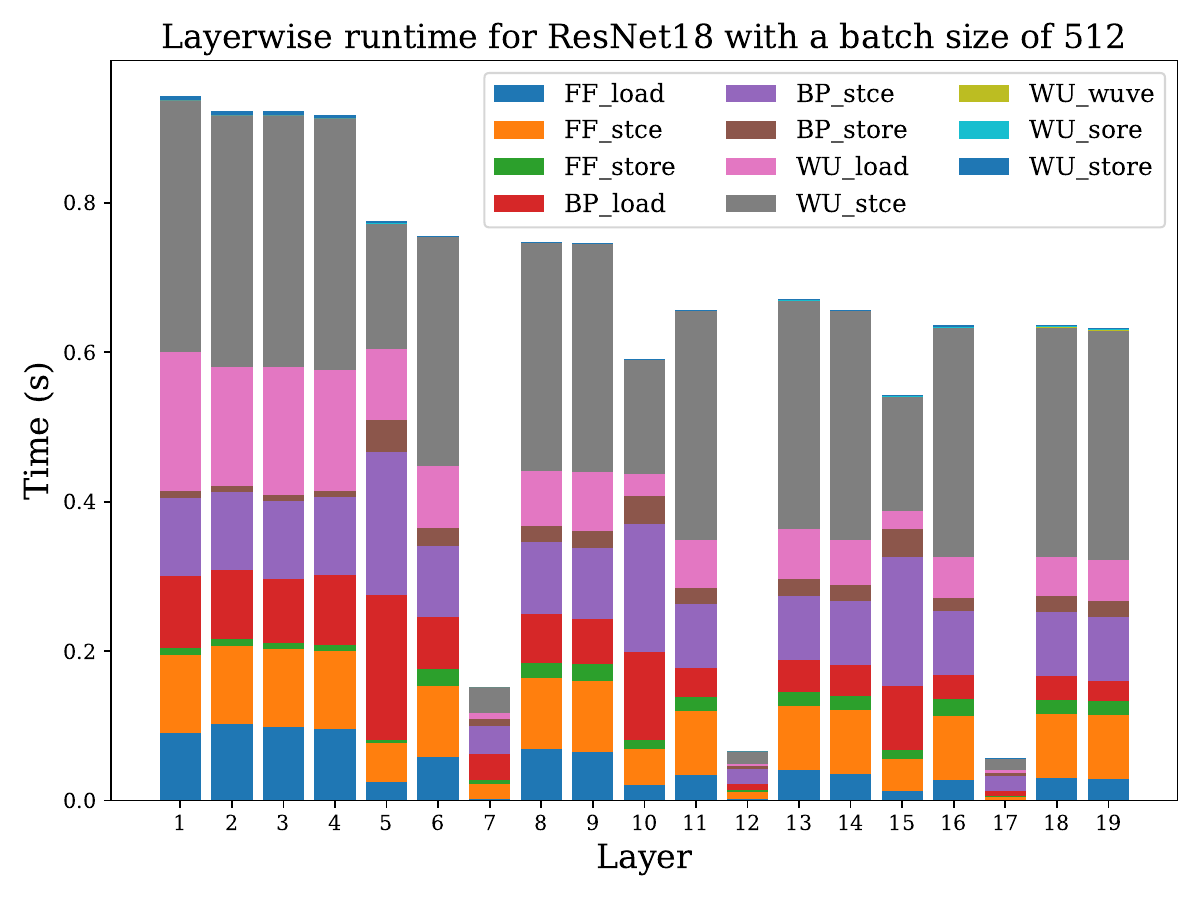}
    \caption{\rTwo{Layer-wise running time per batch of 2:8 sparse training with BDWP in ResNet18 on Tiny ImageNet with a batch size of 512.}} 
	\label{fig:layerwise_profiler}
\end{figure}

To provide a clear picture of the runtime breakdown, Fig.~\ref{fig:layerwise_profiler} presents the running time per batch of BDWP sparse training for each \textit{N:M} sparse convolutional layer in ResNet18 on Tiny ImageNet with a batch size of 512.
Note that we purposely did not overlap the memory access and computing cores during our analysis. 
In actual deployment, the running time of memory access and computing can be significantly reduced through the use of double buffering techniques.
In Fig.~\ref{fig:layerwise_profiler}, STCE's running time for FF and BP, which enable 2:8 sparse computing, is significantly lower compared to that for WU, by approximately a quarter of that required for dense computing.
Additionally, WUVE and SORE exhibit a low activation frequency and short latency in task completion, respectively, consuming only a negligible fraction of the total running time.
Overall, the SAT accelerator shows promise in enabling highly computation-efficient DNN training, with SORE's low resource consumption and running time overhead, coupled with significantly reduced number of operations through \textit{N:M} sparse patterns in FF and BP.

\begin{table}[htbp]
\centering
\caption{\rTwo{Performance Comparison of SAT Versus CPU and GPU}}
\label{tab:cmp_cpu_gpu}
\resizebox{0.48\textwidth}{!}{%
\begin{tabular}{@{}ccccc@{}}
\toprule
                                                                                  & \textbf{CPU}                                             & \textbf{GPU}                                                 & \textbf{GPU}                                                 & \textbf{FPGA}                                                        \\ \midrule
\textbf{Platform}                                                                 & \begin{tabular}[c]{@{}c@{}}Intel\\ i9-9900X\end{tabular} & \begin{tabular}[c]{@{}c@{}}\rTwo{NVIDIA}\\ \rTwo{Jetson Nano}\end{tabular} & \begin{tabular}[c]{@{}c@{}}NVIDIA\\ RTX 2080 Ti\end{tabular} & \begin{tabular}[c]{@{}c@{}}Xilinx\\ XCVU9P\end{tabular}              \\ \midrule
\textbf{Frequency}                                                                & 3.50 GHz                                                 & \rTwo{921 MHz}                                                      & 1.35 GHz                                                     & 200 MHz                                                              \\ \midrule
\textbf{Model}                                                                    & \multicolumn{4}{c}{ResNet18 (Batch Size = 512)}                                                                                                                                                                                                                                  \\ \midrule
\textbf{Precision}                                                                & FP32                                                     & \rTwo{FP32+FP16}                                                         & FP32+FP16                                                    & FP32+FP16                                                            \\ \midrule
\textbf{\begin{tabular}[c]{@{}c@{}}Bandwidth\\ (GB/s)\end{tabular}}      & 57.6                                                     & \rTwo{25.6}                                                         & 616                                                          & 25.6                                                                 \\ \midrule
\textbf{Latency (s)}                                                              & 12.91                                                    & \rTwo{61.28}                                                        & 1.72                                                         & 11.98                                                                \\ \midrule
\textbf{\begin{tabular}[c]{@{}c@{}}Power\\ (W)\end{tabular}}                                                                & 165.00                                                   & \rTwo{7.54}                                                         & 238.36                                                       & \begin{tabular}[c]{@{}c@{}}\rTwo{22.38 (avg.)}
\\ \rTwo{20.73 (dense)}\\ \rTwo{24.15 (2:8 sparse)}\end{tabular}                                                               \\ \midrule
\textbf{\begin{tabular}[c]{@{}c@{}}Peak\\ Throughput\\ (GFLOPS)\end{tabular}}     & 2240                                                     & \rTwo{472}                                                          & 76000                                                        & \begin{tabular}[c]{@{}c@{}}409.6 (dense)\\ 1638.4 (2:8 sparse)\end{tabular} \\ \midrule
\textbf{\begin{tabular}[c]{@{}c@{}}Runtime\\ Throughput\\ (GFLOPS)\end{tabular}}  & 423.69                                                   & \rTwo{94.66}                                                        & 3372.52                                                      & \begin{tabular}[c]{@{}c@{}}\rTwo{484.21 (avg.)}\\\rTwo{280.31 (dense)}\\ \rTwo{702.54 (2:8 sparse)}\end{tabular}                                                               \\ \midrule
\textbf{\begin{tabular}[c]{@{}c@{}}Energy\\ Efficiency\\ (GFLOPS/W)\end{tabular}} & 2.57                                                     & \rTwo{12.56}                                                        & 14.15                                                        & \begin{tabular}[c]{@{}c@{}}\rTwo{21.64 (avg.)}\\\rTwo{13.52 (dense)}\\ \rTwo{29.09 (2:8 sparse)}\end{tabular}                                                                 \\ \bottomrule
\end{tabular}%
}
\end{table}

\subsection{Comparison with CPU and GPU}

To evaluate the potential of SAT for efficient DNN training, we compare it against CPU and GPU platforms. 
The CPU baseline is the Intel Core i9-9900X processor, equipped with 19.25 MB L3 cache, 10 physical cores, 20 threads running at 3.50 GHz, and a thermal design power (TDP) of 165 W. 
\rTwo{Meanwhile, the GPU baselines are the NVIDIA RTX 2080 Ti card and Jetson Nano.}
\rTwo{The former achieves a peak throughput of 76 TFLOPS equipped with 4352 CUDA cores running at 1.35 GHz and a TDP of 250 W, and the latter is a competitive candidate for energy-efficient edge computing scenarios with a peak throughput of 472 GFLOPS.}
As shown in Table \ref{tab:cmp_cpu_gpu}, 
the peak throughput of SAT can achieve 409.6 GOPs for dense operations, which closely aligns with Jetson Nano’s performance, and achieve 1638.4 GOPs for 2:8 sparse operations. 

\rTwo{Energy efficiency across CPU, GPUs, and SAT is evaluated to highlight the potential in \textit{N:M} sparse training.
The Intel performance counter monitor utility \cite{pcm} is used to measure the actual CPU power consumption. We use \textit{nvidia-smi} on RTX 2080 Ti and \textit{jtop} on Jetson Nano to measure the GPU run-time power.}
For fair comparison against SAT, we use PyTorch v1.10 to perform convolutional layers that have been arranged in MatMul form in ResNet18 on both CPU and GPU with a batch size of 512.
As presented in Fig.~\ref{tab:cmp_cpu_gpu}, SAT achieves 8.42$\times$ energy efficiency improvement compared to the CPU baseline. 
\rTwo{Moreover, compared to Jetson Nano and RTX 2080 Ti, SAT improves energy efficiency by 1.72$\times$ and 1.53$\times$, respectively.}

\begin{figure} [tbp] 
	\centering
	\includegraphics[width=0.5\textwidth]{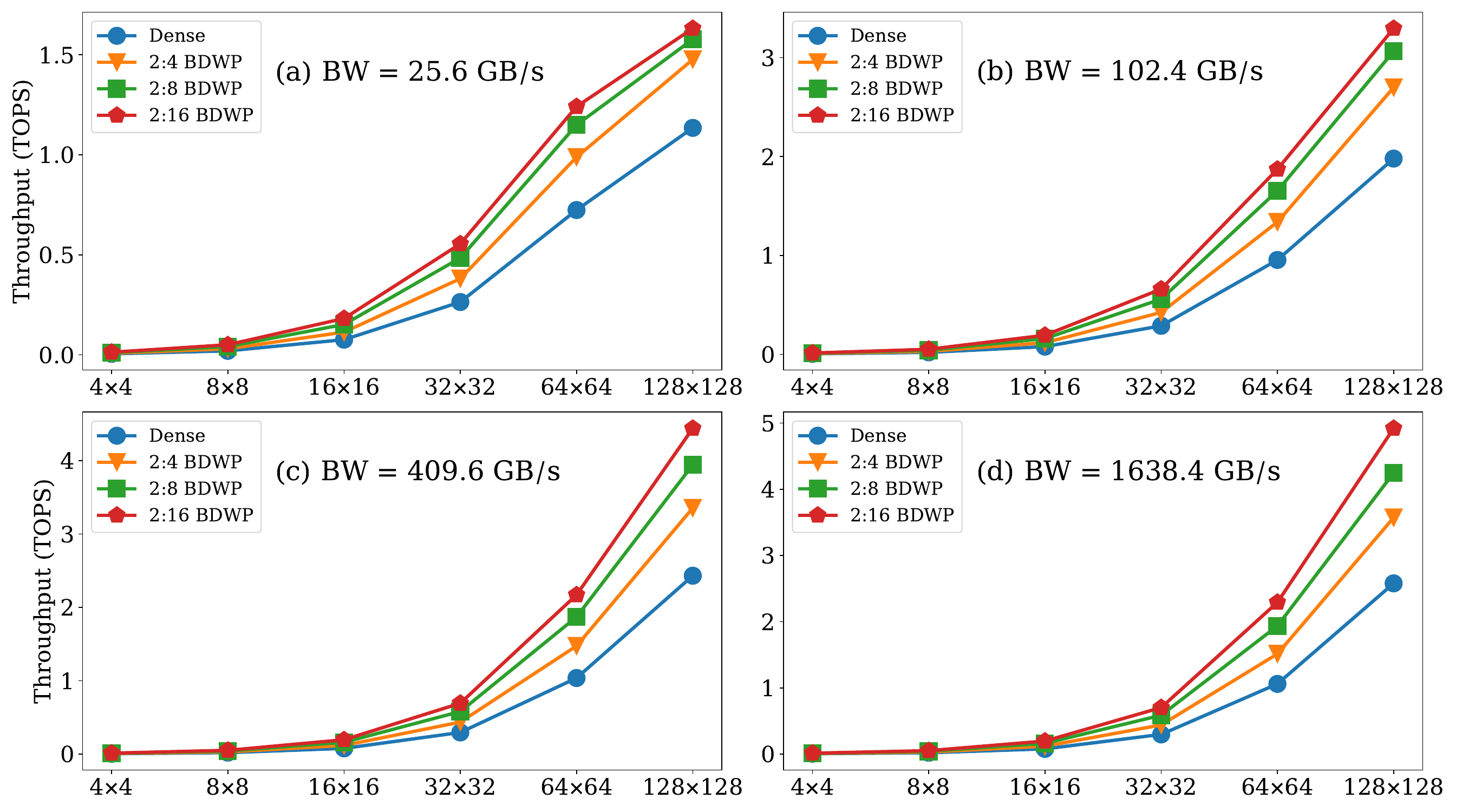}
    \caption{Runtime training throughput of ResNet18 with different available off-chip memory bandwidths when scaling the number of USPEs in STCE.}
	\label{fig:res_scale}
\end{figure}

\rTwo{To make a fair comparison with the RTX 2080 Ti, we scale SAT by changing the number of USPEs in STCE and the off-chip bandwidth, while keeping other factors unchanged.}
\rTwo{The experimental results are shown in Fig.~\ref{fig:res_scale}, where the X-axis represents the systolic array size of STCE in SAT. It can be seen that the number of USPEs and the off-chip bandwidth have a significant impact on the runtime throughput of training. When the off-chip bandwidth is 409.6 GB/s, as shown in Fig.~\ref{fig:res_scale}~(c), which is less than the 616 GB/s of the RTX 2080 Ti GPU as shown in Table~\ref{tab:cmp_cpu_gpu}, the runtime throughput of SAT executing 2:8 BDWP reaches 3.9 TOPS, which is greater than the runtime throughput of the RTX 2080 Ti GPU in training ResNet18 (only 3.4 TOPS). In addition, the dense peak performance of 2:8 SAT under this configuration is 6.6 TOPS, and the sparse peak performance of 2:8 SAT is 26.2 TOPS, which is also significantly less than the 76 TOPS of the RTX 2080 Ti GPU. This shows that SAT after scaling has a higher computational utilization for training a ResNet18.}

\begin{table*}[htbp]
\centering
\caption{Comparison of Prior FPGA-based Training Accelerators}
\label{tab:cmp_accel}
\resizebox{\textwidth}{!}{%
\begin{tabular}{@{}cccccccccc@{}}
\toprule
\textbf{Accelerator}                        & \textbf{Platform} & \textbf{Network}   & \textbf{Precision} & \textbf{DSP Util.} & \textbf{\begin{tabular}[c]{@{}c@{}}Freq. \\ (MHz)\end{tabular}} & \textbf{\begin{tabular}[c]{@{}c@{}}Power \\ (W)\end{tabular}} & \textbf{\begin{tabular}[c]{@{}c@{}}Throughput\\ (GOPS)\end{tabular}} & \textbf{\begin{tabular}[c]{@{}c@{}}Comp. Effi.\\ (GOPS/DSP)\end{tabular}} & \textbf{\begin{tabular}[c]{@{}c@{}}Energy Effi.\\ (GOPS/W)\end{tabular}} \\ \midrule
\textbf{SAT (this work)}                    & \textbf{XCVU9P}   & \textbf{ResNet-18} & \textbf{FP16+FP32} & \textbf{1228}               & \textbf{200}                                                    & \textbf{22.38}                                                & \textbf{484.21}                                                      & \textbf{0.39}                                                             & \textbf{21.64}                                                           \\
TODAES'22 \cite{tang2022ef}                 & ZCU102            & VGG-16             & FP32               & 1508               & 100                                                             & 7.71                                                          & 46.99                                                                & 0.03                                                                      & 6.09                                                                     \\
FPGA'20 \cite{he2020fecaffe}                & Stratix 10        & AlexNet            & FP32               & 1796               & 253                                                             & N/A                                                           & $\sim$24.00                                                               & 0.01                                                                      & N/A                                                                      \\
FPT'17 \cite{liu2017fpga}                   & ZU19EG            & LeNet-10           & FP32               & 1500               & 200                                                             & 14.24                                                         & 86.12                                                                & 0.06                                                                      & 6.05                                                                     \\
ICCAD'20 \cite{venkataramanaiah2020fpga}    & Stratix 10 MX     & VGG-like           & FP16               & 1046               & 185                                                             & $\sim$20.00                                                        & $\sim$158.54                                                              & 0.15                                                                      & $\sim$9.00                                                                    \\
OJCAS'23 \cite{tsai2023chip}                & ZCU104            & AlexNet            & BFP16              & 1285               & 200                                                             & 6.44                                                          & 102.43                                                                & 0.08                                                                      & 15.90                                                                    \\
AICAS'21 \cite{chen2021eile}                & XC7Z100           & FC                 & INT16              & 64                 & 150                                                             & 2.50                                                          & 19.20                                                                & 0.30                                                                      & 7.68                                                                     \\
FPL'19 \cite{venkataramanaiah2019automatic} & Stratix 10 GX     & VGG-like           & INT16              & 1699               & 240                                                             & 20.60                                                         & 163.00                                                               & 0.09                                                                      & 7.90                                                                     \\ \midrule
FPL'19 \cite{nakahara2019fpga}                           & XCVU9P            & AlexNet            & FP9                & 1106               & 200                                                             & 75.00                                                         & 375.61                                                               & 0.34                                                                      & 5.00                                                                     \\
ISVLSI'21 \cite{shao2021fpga}               & VC709             & VGG-like           & INT8               & 2324               & 200                                                             & 16.27                                                         & 771.00                                                               & 0.33                                                                      & 47.38                                                                    \\
JOS'20 \cite{luo2020towards}                & XCVU9P            & VGG-like           & INT8               & 4202               & 200                                                             & 13.50                                                         & 1417.00                                                              & 0.34                                                                      & 104.96                                                                   \\
TNNLS'22 \cite{lu2022eta}                   & VC709             & VGG-16             & PINT8              & 1728               & 200                                                             & 8.44                                                          & 610.98                                                               & 0.35                                                                      & 72.37                                                                    \\ \bottomrule
\end{tabular}%
}
\end{table*}

\subsection{Comparison with FPGA-based Training Accelerators}

Table~\ref{tab:cmp_accel} presents a comparison between SAT and existing state-of-the-art FPGA-based training accelerators for DNNs. 
Our SAT outperforms other architectures equipped with FP16 or higher numerical formats, exhibiting superior performance in terms of throughput, computational efficiency, and energy efficiency. 
Specifically, SAT improves the training throughput by 2.97$\sim$25.22$\times$, computational efficiency by 1.3$\sim$39$\times$, and energy efficiency by 1.36$\sim$3.58$\times$ when compared to \cite{venkataramanaiah2020fpga, tang2022ef, he2020fecaffe, liu2017fpga, venkataramanaiah2019automatic, tsai2023chip, chen2021eile}.
The superior results of SAT can be attributed to its efficient hardware implementation of \textit{N:M} sparsity acceleration. 
\rTwo{First, by exploiting the parallelism offered by large batch sizes, SAT can significantly increase the throughput of the training process, which is a significant improvement over prior FPGA-based accelerators \cite{venkataramanaiah2020fpga, venkataramanaiah2019automatic} that use small batch sizes for DNN training.
Second, a more efficient dataflow design is adopted to efficiently cover the data loading and computation process, leading to high throughput and energy efficiency of SAT.
Third, we exploit 2:8 sparsity in the forward and backward training processes, which significantly reduces the number of training operations by 48\% on average, leading to improvements in throughput and energy efficiency.}
By leveraging the potential of \textit{N:M} sparsity acceleration on hardware, SAT presents a novel approach to efficient DNN training that is orthogonal \cite{park2022quantized} to prior works that employ reduced numerical precision \cite{nakahara2019fpga, shao2021fpga, lu2022eta, luo2020towards}. 
Our results demonstrate that SAT is a promising FPGA-based accelerator for DNN training that significantly outperforms existing state-of-the-art solutions, highlighting the potential of \textit{N:M} sparse acceleration for efficient DNN training.

\subsection{Discussion}

The effectiveness of \textit{N:M} sparse DNN training has been demonstrated both at the algorithm and hardware levels.
From an algorithm perspective, BDWP achieves significant computational reduction without sacrificing model accuracy compared to other state-of-the-art \textit{N:M} sparse training methods.
Moreover, BDWP can be easily integrated with AMP training pipeline. 
From a hardware perspective, SAT is efficient in supporting \textit{N:M} sparse DNN training.
SORE incurs less than 1\% hardware overhead, and STCE supports flexible dataflows, as well as both regular dense and computation-efficient \textit{N:M} sparse operations, significantly improving DNN training efficiency.
Deployment of BDWP on SAT reduces training time by 43\%, resulting in 1.75$\times$ training acceleration compared to dense training.
Additionally, SAT outperforms CPU and GPU in energy efficiency and also shows significant improvements over prior state-of-the-art FPGA-based training accelerators.
These results demonstrate the proposed \textit{N:M} sparse training scheme is particularly promising for achieving efficient and rapid training for increasingly large DNN models.

\section{Conclusion} \label{sec:concls}

In this paper, we present an efficient \textit{N:M} sparse DNN training scheme on FPGA exploiting optimizations of algorithm, architecture, and dataflow aspects.
At the algorithm level, a novel bidirectional weight pruning method, dubbed BDWP, is first proposed to significantly reduce the number of operations while maintaining model accuracy.
At the architecture level, a sparse accelerator for DNN training, namely SAT, is further developed to support computation-efficient \textit{N:M} sparse operations besides the regular dense operations efficiently.
At the dataflow level, multiple optimization techniques further increase hardware utilization, improving the throughput of SAT.
Experimental results show our \textit{N:M} sparse training scheme can significantly improve the training throughput by 2.97$\sim$25.22$\times$ and the energy efficiency by 1.36$\sim$3.58$\times$ compared to state-of-the-art FPGA training accelerators.
As the computations involved in DNN training are rapidly increasing, this work should be helpful for developing efficient sparse DNN training.

\section*{Acknowledgment}

\rTwo{The authors would like to sincerely thank their reviewers for the valuable feedback.}
\ifCLASSOPTIONcaptionsoff
  \newpage
\fi



\bibliographystyle{IEEEtran}
\bibliography{IEEEabrv, ref_full}
%



\end{document}